\documentclass{article}

\usepackage{arxiv}
\usepackage[utf8]{inputenc} % allow utf-8 input
\usepackage[T1]{fontenc}    % use 8-bit T1 fonts
\usepackage{hyperref}       % hyperlinks
\usepackage{url}            % simple URL typesetting
\usepackage{booktabs}       % professional-quality tables
\usepackage{amsmath}
\usepackage{amssymb}
\usepackage{amsfonts}       % blackboard math symbols
\usepackage{nicefrac}       % compact symbols for 1/2, etc.
\usepackage{microtype}      % microtypography
\usepackage{natbib}
\usepackage{siunitx}
\usepackage{mathtools}

\newcommand{\scale}{0.8}

\title{Using machine learning to correct  model error in data assimilation and forecast applications}

\author{
Alban Farchi\\
CEREA \\
joint laboratory École des Ponts ParisTech and EDF R\&D\\
Champs-sur-Marne, France\\
\texttt{alban.farchi@enpc.fr}\\
\AND
Patrick Laloyaux\\
ECMWF\\
Shinfield Park\\
Reading, United Kingdom\\
\AND
Massimo Bonavita\\
ECMWF\\
Shinfield Park\\
Reading, United Kingdom\\
\AND
Marc Bocquet\\
CEREA\\
joint laboratory École des Ponts ParisTech and EDF R\&D\\
Champs-sur-Marne, France
}

\begin{document}

\maketitle

\begin{abstract}
The idea of using machine learning (ML) methods to reconstruct the dynamics of a system is the topic of recent studies in the geosciences, in which the key output is a surrogate model meant to emulate the dynamical model. In order to treat sparse and noisy observations in a rigorous way, ML can be combined to data assimilation (DA). This yields a class of iterative methods in which, at each iteration a DA step assimilates the observations, and alternates with a ML step to learn the underlying dynamics of the DA analysis. In this article, we propose to use this method to correct the error of an existent, knowledge-based model. In practice, the resulting surrogate model is a hybrid model between the original (knowledge-based) model and the ML model. We demonstrate numerically the feasibility of the method using a two-layer, two-dimensional quasi-geostrophic channel model. Model error is introduced by the means of perturbed parameters. The DA step is performed using the strong-constraint 4D-Var algorithm, while the ML step is performed using deep learning tools. The ML models are able to learn a substantial part of the model error and the resulting hybrid surrogate models produce better short- to mid-range forecasts. Furthermore, using the hybrid surrogate models for DA yields a significantly better analysis than using the original model.
\end{abstract}


%--------------------------------------------------
\section{Introduction}
%--------------------------------------------------

The recent and remarkable emergence of machine learning (ML) methods, and in particular deep learning (DL), can be explained by several factors, among which (i) the increasing computational capabilities, (ii) the access to large datasets for training, and (iii) the development of efficient and user-friendly libraries \citep{lecun-2015, goodfellow-2016, chollet-2018}. Impressive results have been obtained in a wide range of problems using DL to the extent that DL has become state-of-the-art for many different applications: computer vision, natural language processing, signal processing, etc...

In numerical weather prediction, even if the physical laws governing the system dynamics are reasonably well known, the numerical models are affected by errors. This model error could come, for example, from misrepresented physical processes or from unresolved small-scale processes. It is legitimate to wonder whether ML methods can make use of the large amount of Earth observations, available in particular through remote sensing, to try and provide better numerical models. This question is the topic of numerous studies in the geosciences, whose objective is to construct a dynamical model using only observations of a physical system. The output dynamical model is often called the surrogate model to emphasise its difference with the (true) dynamical model. Several ML methods have already been tested to construct a surrogate model for low-order chaotic systems. Examples include the analogs \citep{lguensat-2017}, recurrent neural networks \citep{park-1994, park-2010}, residual neural networks to represent either the resolvent \citep{brajard-2020, dueben-2018, scher-2019} or the underlying ordinary differential equations \citep[ODEs,][]{fablet-2018, long-2018, bocquet-2019a, bocquet-2020}, and reservoir computing \citep{pathak-2017, pathak-2018, pathak-2018a}.

In all cases, the learning step is a variational problem which consists in optimising a loss function expressing the discrepancy between the surrogate model prediction and the actual predictions (\textit{e.g.} the observations). This process is called supervised learning, and it requires an accurate knowledge of the true state of the system, which is why in most studies it is assumed that the system is fully observed without noise. Adding a small observation noise is possible and acts like regularisation, but currently, standard ML methods alone cannot handle sparse observations. Yet, \citet{brajard-2020} have shown how ML can be combined with data assimilation (DA) to overcome this issue. In their method, DA and ML steps are alternated to produce estimates of both the state and the surrogate model with an increasing accuracy. This idea has then been framed into a unifying Bayesian formalism by \citet{bocquet-2020}.

However, the application of such algorithms to realistic models is not immediate, first because the computational cost of repeated DA steps may be prohibitive, and second because the surrogate model initialisation is critical: a cold start could lead to divergence from the true evolution of the system. Therefore, instead of constructing the surrogate model from scratch, we could build a hybrid model on top of an already existent, knowledge-based model as proposed, \textit{e.g.}, by \citet{pathak-2018a, watson-2019, brajard-2020b}. In practice, the trainable part of the surrogate model would be correcting the error of the knowledge-based model: this is an offline model error estimation method. As explained by \citet{jia-2019, watson-2019}, error-correcting learning problems are in general easier in the sense that they require smaller ML models and less training data.

In parallel, there have been recent efforts in the research and operational DA communities to develop weak-constraint DA methods, \textit{i.e.} relaxing the perfect model assumption. The main difficulty in this context is that a model trajectory is not uniquely described by the initial condition, which significantly increases the dimension of the problem. This effort has lead, for example, to the development of the iterative ensemble Kalman filter in presence of additive noise (IEnKF-Q) by \citet{sakov-2018}, and of the forcing formulation of weak-constraint 4D-Var by \citet{laloyaux-2020}, and its implementation in the operational framework developed at European Centre for Medium-Range Weather Forecasts (ECMWF) \citep{laloyaux-2020b}. Both algorithms are online model error estimation methods, and could provide baseline of comparison when using a trained surrogate model for DA.

The aim of this paper is to investigate the possibility of learning the error of a knowledge-based model from a database of sparse and noisy observations using standard DL methods. The experiments are performed using a two-layer quasi-geostrophic channel model implemented in the ECMWF Object Oriented Prediction System - Integrated Forecast System \citep[ECMWF OOPS-IFS,][]{bonavita-2017}. The model yields reasonably complex problems (two-dimensional, two layers, 1600 variables in total) while being sufficiently small to allow extensive tests in a controlled environment. Furthermore, it has already been used to validate the weak-constraint 4D-Var algorithm by \citet{laloyaux-2020}. 

The article is organised as follows. Section~\ref{sec:sec2-methodology} introduces the main methodological aspects of the study. Section~\ref{sec:sec3-qg-model} presents the dynamical model used in the numerical experiments. Sections~\ref{sec:sec4-original-da-step} to \ref{sec:sec6-corrected-da-step} illustrate the method: first, the original DA step in Section~\ref{sec:sec4-original-da-step}, then the ML step in Section~\ref{sec:sec5-ml-step}, and finally the corrected DA step in Section~\ref{sec:sec6-corrected-da-step}. Conclusions are drawn in Section~\ref{sec:sec7-conclusions}.

%--------------------------------------------------
\section{Methodological aspects}
\label{sec:sec2-methodology}
%--------------------------------------------------

%--------------------------------------------------
\subsection{A Bayesian framework for machine learning and data assimilation}
%--------------------------------------------------

We follow the system state $\mathbf{x}_{k}\in\mathbb{R}^{N_{\mathsf{x}}}$ at discrete times $t_{k}, k\in\mathbb{N}$. Suppose that the evolution of the state is governed by
\begin{equation}
    \mathbf{x}_{k+1} = \mathcal{M}^{\mathsf{t}}_{k}\left(\mathbf{x}_{k}\right),
\end{equation}
where $\mathcal{M}^{\mathsf{t}}_{k}:\mathbb{R}^{N_{\mathsf{x}}}\to\mathbb{R}^{N_{\mathrm{x}}}$ is the resolvent of the \emph{unknown} true dynamical model from $t_{k}$ to $t_{k+1}$. Our objective is to provide a surrogate to the true model, whose resolvent can then be used to predict $\mathbf{x}_{k+1}$ given $\mathbf{x}_{k}$. For simplicity, we make the additional assumption that $\Delta t=t_{k+1}-t_{k}$ is constant.

A standard ML approach to this problem consists in minimising the cost function 
\begin{equation}
    \label{eq:sec2-ml-cost}
    \mathcal{J}\left(\mathbf{p}\right) \triangleq 
        \frac{1}{2} \sum_{k=0}^{N_{\mathsf{t}}-1} 
        \left\| \mathbf{x}_{k+1} - \mathcal{M}_{k}\left(\mathbf{p}, \mathbf{x}_{k}\right) \right\|^{2}_{\mathbf{Q}^{-1}_{k}} 
        + \mathcal{L}\left(\mathbf{p}\right),
\end{equation}
where $N_{\mathsf{t}}$ is the length of the training trajectory, $\mathbf{x}\mapsto\mathcal{M}_{k}\left(\mathbf{p}, \mathbf{x}\right)$ is the resolvent of the surrogate model from $t_{k}$ to $t_{k+1}$, $\mathbf{p}\in\mathbb{R}^{N_{\mathsf{p}}}$ is the set of coefficients used to define the surrogate model (for instance the weights and biases of a neural network), and $\mathcal{L}$ is a regularisation term. Minimising Equation~\eqref{eq:sec2-ml-cost} amounts to finding the surrogate model which best fits the trajectory (supervised learning).

The main drawback of this approach is that, for realistic applications, the true state of the system $\mathbf{x}_{k}$ is only known through the observation vectors $\mathbf{y}_{k}\in\mathbb{R}^{N_{\mathrm{y}}}$. The observations are related to the system state by
\begin{equation}
    \mathbf{y}_{k} = \mathcal{H}_{k}\left(\mathbf{x}_{k}\right) + \mathbf{v}_{k},
\end{equation}
where $\mathcal{H}_{k}:\mathbb{R}^{N_{\mathrm{x}}}\to\mathbb{R}^{N_{\mathrm{y}}}$ is the observation operator at time $t_{k}$ and $\mathbf{v}_{k}$ is the (random) observation noise. Substituting $\mathbf{x}_{k}$ for $\mathbf{y}_{k}$ in Equation~\eqref{eq:sec2-ml-cost} is possible if the observations are full, \textit{i.e.} if $\mathcal{H}_{k}=\mathbf{I}$, the identity operator. In this case, the observation noise $\mathbf{v}_{k}$ would act as a regularisation of the learning phase \citep{bishop-1995}. However, in order to process the general case in which observations can be sparse, Equation~\eqref{eq:sec2-ml-cost} needs to be expanded to the more general cost function \citep{hsieh-1998, abarbanel-2018, bocquet-2019a, bocquet-2020}
\begin{equation}
    \label{eq:sec2-daml-cost}
    \mathcal{J}\left(\mathbf{p}, \mathbf{x}_{0:N_{\mathsf{t}}}\right) \triangleq
    \frac{1}{2} \left\| \mathbf{x}_{0} - \mathbf{x}^{\mathsf{b}}_{0} \right\|^{2}_{\mathbf{B}^{-1}}
    +\frac{1}{2} \sum_{k=0}^{N_{\mathsf{t}}}
    \left\| \mathbf{y}_{k} - \mathcal{H}_{k}\left(\mathbf{x}_{k}\right) \right\|^{2}_{\mathbf{R}^{-1}_{k}}
    + \frac{1}{2} \sum_{k=0}^{N_{\mathsf{t}}-1} 
    \left\| \mathbf{x}_{k+1} - \mathcal{M}_{k}\left(\mathbf{p}, \mathbf{x}_{k}\right) \right\|^{2}_{\mathbf{Q}^{-1}_{k}}
    + \mathcal{L}\left(\mathbf{p}\right),
\end{equation}
where $\mathbf{x}_{0:N_{\mathsf{t}}}$ is the system trajectory $\left\{\mathbf{x}_{k},\; k=0\ldots N_{\mathsf{t}}\right\}$, the matrix norm notation $\left\|\mathbf{v}\right\|^{2}_{\mathbf{M}}$ stands for $\mathbf{v^{\top}Mv}$, and $\mathbf{x}^{\mathsf{b}}_{0}$ is the background state, \textit{i.e.} the prior estimate of the initial state. Furthermore, in Equation~\eqref{eq:sec2-daml-cost} it is implicitly assumed that:
\begin{itemize}
    \item the background error $\mathbf{x}_{0} - \mathbf{x}^{\mathsf{b}}_{0}$ follows a zero-mean Gaussian distribution with covariance matrix $\mathbf{B}$;
    \item the observation errors follow zero-mean Gaussian distributions with covariance matrices $\mathbf{R}_{0:N_{\mathsf{t}}}$;
    \item the model errors follow zero-mean Gaussian distributions with covariance matrices $\mathbf{Q}_{0:N_{\mathsf{t}}-1}$;
    \item the model and observation errors are both uncorrelated in time;
    \item the model and observation errors are uncorrelated.
\end{itemize}
Equation~\eqref{eq:sec2-daml-cost} is very similar to a typical weak-constraint 4D-Var cost function \citep{tremolet-2006}. In Equation~\eqref{eq:sec2-daml-cost}, the system trajectory $\mathbf{x}_{0:N_{\mathsf{t}}}$ is unknown. This is not the case in Equation~\eqref{eq:sec2-ml-cost}, which is why the background term is missing in Equation~\eqref{eq:sec2-ml-cost}. Finally, in the case where the observations are full ($\mathcal{H}_{k}=\mathbf{I}$) and noiseless ($\mathcal{R}_{k}=\mathbf{0}$), Equations~\eqref{eq:sec2-ml-cost} and \eqref{eq:sec2-daml-cost} are equivalent.

%--------------------------------------------------
\subsection{Optimisation strategy}
\label{ssec:sec2-optimisation-strategy}
%--------------------------------------------------

If possible, one can try and minimise Equation~\eqref{eq:sec2-daml-cost} for both the parameters and the state trajectory at the same time. This method has been used, for example, by \citet{bocquet-2019a} to reconstruct the dynamics of low-order models with as few parameters as possible. In their method, they chose to implement the minimisation step using the L-BFGS algorithm \citep{byrd-1995}. However for realistic problems, a joint minimisation may be very difficult to implement, because the state space $\mathbb{R}^{N_{\mathsf{x}}}$ is high-dimensional, and because in order to get an accurate description of the dynamics, the number of parameters $N_{\mathsf{p}}$ and the length of the training trajectory $N_{\mathsf{t}}$ must be large enough. 

\begin{figure*}[tb]
    \centering
    \includegraphics[scale=\scale]{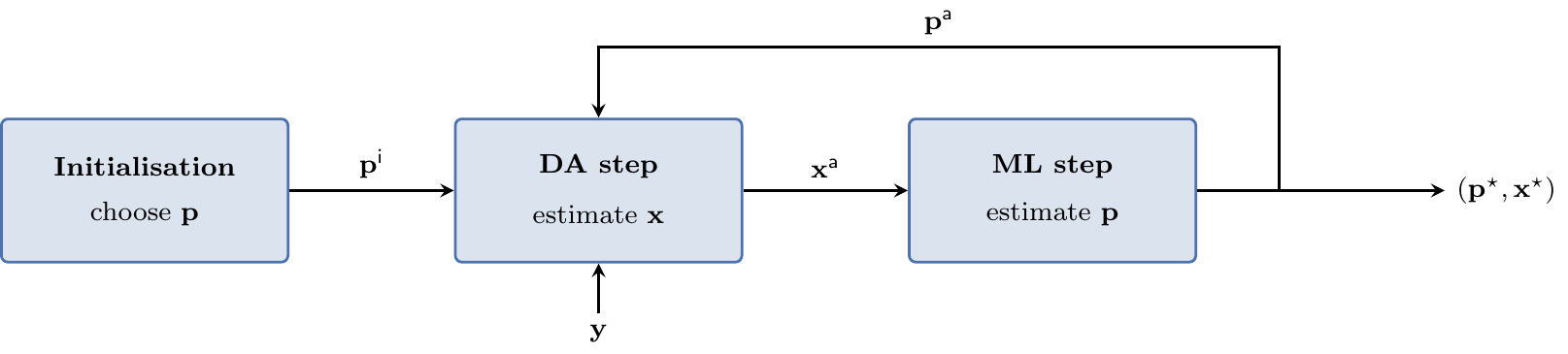}
    \caption{Minimisation strategy for Equation~\eqref{eq:sec2-daml-cost}: alternate DA steps with ML steps to estimate the parameters $\mathbf{p}$ and the system trajectory $\mathsf{x}_{0:N_{\mathsf{t}}}$ with an increasing accuracy.}
    \label{fig:sec2-daml-method}
\end{figure*}

Recognising that the parameters $\mathbf{p}$ and the system trajectory $\mathbf{x}_{0:N_{\mathsf{t}}}$ are of different nature, \citet{bocquet-2020} suggest that a coordinate descent could be more efficient. In this case, the strategy would be to alternate a standard DA step to estimate the system trajectory $\mathbf{x}_{0:N_{\mathsf{t}}}$ with a standard ML step to estimate the parameters $\mathbf{p}$. This method, illustrated in Figure~\ref{fig:sec2-daml-method}, is highly flexible because both DA and ML steps are totally independent. Furthermore, the use of ML in this framework is more technical than conceptual: the formalism is that of DA. This method has been used, for example, by \citet{brajard-2020, bocquet-2020} to reconstruct the dynamics of low-order models using convolutional neural networks. Yet, the application to a realistic model is not immediate for two reasons. First, the initialisation (the choice of the initial set of parameters $\mathbf{p}^{\mathsf{i}}$) is critical and for a complex model, a cold start could yield numerical instability and quick divergence. Second, the number of DA--ML cycles required until convergence can be quite high. For example, the application of \citet{brajard-2020} to a low-order model needs several dozen of cycles to converge. By contrast, a single DA step with a realistic model can be very costly, especially if the training trajectory length $N_{\mathsf{t}}$ is high.

A natural idea to overcome these issues is, instead of constructing the surrogate model from scratch, to build a hybrid model using an already existent, knowledge-based model \citep{pathak-2018a, brajard-2020b}. In fact, this would mean that we would try and correct the error of the knowledge-based model. In the absence of any prior information on the error of the original (knowledge-based) model, the initialisation is very simple: the correction is set to zero. In other words, the first DA step is performed with the original model. Furthermore, correcting the original model is likely to be an easier learning problem than constructing the full model dynamics \citep{jia-2019, watson-2019}, in particular if most of the physical processes are described in the original model. This means that we could use smaller ML models and less training data, but also that the number of DA--ML cycles required until convergence could be smaller. 

%--------------------------------------------------
\subsection{A hybrid surrogate model}
%--------------------------------------------------

Two formulations can emerge from the idea of correcting an original model, depending on whether the correction is applied to the resolvent or directly to the tendencies (in other words in the underlying ODEs or partial differential equations).

%--------------------------------------------------
\subsubsection{Correcting the resolvent}
\label{sssec:sec2-correcting-resolvent}
%--------------------------------------------------

If we choose to apply the correction to the resolvent, then the resolvent of the hybrid surrogate model from $t_{k}$ to $t_{k+1}$ can be written
\begin{equation}
    \label{eq:sec2-def-hybrid-resolvent}
    \mathcal{M}_{k}\left(\mathbf{p}, \mathbf{x}\right) \triangleq
    \mathcal{M}^{\mathsf{o}}_{k}\left(\mathbf{x}\right)
    + \mathcal{M}^{\mathsf{ml}}_{k}\left(\mathbf{p}, \mathbf{x}\right),
\end{equation}
where $\mathcal{M}^{\mathsf{o}}_{k}$ is the resolvent of the original model and $\mathcal{M}^{\mathsf{ml}}_{k}$ is the ML model. It is important to realise here that the correction has been added to the original resolvent. In particular, it means that the output of the ML model lies in the state space. This choice has the advantage of being simple from a methodological point of view while having the potential to handle any model error. However, other choices are possible, for example multiplying the original resolvent by the correction term or even composing the original resolvent and the ML model. The latter choices would be more appropriate if the true model error has a \emph{simpler} definition in multiplicative form $\mathcal{M}^{\mathsf{t}}/\mathcal{M}^{\mathsf{o}}$ or in composition form $\mathcal{M}^{\mathsf{t}}\circ\left(\mathcal{M}^{\mathsf{o}}\right)^{-1}$ (provided that these quantities exist) than in additive form $\mathcal{M}^{\mathsf{t}}-\mathcal{M}^{\mathsf{o}}$.

Substituting Equation~\eqref{eq:sec2-def-hybrid-resolvent} into the cost function, Equation~\eqref{eq:sec2-daml-cost}, yields
\begin{equation}
    \mathcal{J}\left(\mathbf{p}, \mathbf{x}_{0:N_{\mathsf{t}}}\right) = 
    \frac{1}{2} \left\| \mathbf{x}_{0} - \mathbf{x}^{\mathsf{b}}_{0} \right\|^{2}_{\mathbf{B}^{-1}}
    +\frac{1}{2} \sum_{k=0}^{N_{\mathsf{t}}}
    \left\| \mathbf{y}_{k} - \mathcal{H}_{k}\left(\mathbf{x}_{k}\right) \right\|^{2}_{\mathbf{R}^{-1}_{k}}
    + \frac{1}{2} \sum_{k=0}^{N_{\mathsf{t}}-1} 
    \left\| \left\{ \mathbf{x}_{k+1} - \mathcal{M}^{\mathsf{o}}_{k}\left(\mathbf{x}_{k}\right) \right\} - \mathcal{M}^{\mathsf{ml}}_{k}\left(\mathbf{p}, \mathbf{x}_{k}\right) \right\|^{2}_{\mathbf{Q}^{-1}_{k}}
    + \mathcal{L}\left(\mathbf{p}\right).
\end{equation}
Two points should be highlighted. First, the DA steps (except for the first one) are performed using the hybrid model defined by Equation~\eqref{eq:sec2-def-hybrid-resolvent}. If the ML model $\mathcal{M}^{\mathsf{ml}}_{k}$ is implemented using standard ML tools, for instance neural networks (NNs), then there is almost no technical difference with the first DA step, performed using the original model only. In particular, if needed (for example, when using variational DA) the adjoint and tangent linear (TL) of the trainable model can be obtained through the ML library. Second, the ML steps are similar to a standard ML step, the difference being that the ML model has to learn the error of the original model, which has consequences in the preprocessing stage only.

In summary, this method requires only minor modifications to the existing numerical methods, and should therefore be easy to implement. On the other side, the hybrid model defined by  Equation~\eqref{eq:sec2-def-hybrid-resolvent} is limited in the sense that it can only make predictions for a $\Delta t$-multiple time horizon.

%--------------------------------------------------
\subsubsection{Correcting the tendencies}
\label{sssec:sec2-correcting-flow}
%--------------------------------------------------

Suppose now that the correction is directly applied to the tendencies. The surrogate model can then be defined by the following ODE:
\begin{equation}
    \label{eq:sec2-def-hybrid-ode}
    \frac{\mathrm{d}\mathbf{x}}{\mathrm{d}t} = \mathcal{F}^{\mathsf{o}}\left(\mathbf{x}\right) + \mathcal{F}^{\mathsf{ml}}\left(\mathbf{p}, \mathbf{x}\right),
\end{equation}
where $\mathcal{F}^{\mathsf{o}}$ is the tendency of the original model and $\mathcal{F}^{\mathsf{ml}}$ is the ML model. These hybrid tendencies must then be integrated from $t_{k}$ to $t_{k+1}$ to form the resolvent of the surrogate model $\mathcal{M}_{k}$.

This formulation essentially leaves the cost function, Equation~\eqref{eq:sec2-daml-cost}, unchanged. However, the tendencies of the original model and the trainable model become intricate as a result of the integration and this has implications for the variational calculus. First, the contribution of the ML model to the adjoint and TL model of the resolvent of the surrogate model (potentially needed for variational DA) is nonlinear. Second, the gradient of the resolvent of the surrogate model with respect to $\mathbf{p}$ (usually needed for the ML steps) is not trivial, and in particular it may depend on the TL model of the resolvent of the original model.

In summary, and by contrast with the previous formulation, this method may require some substantial modifications to the existing numerical methods, but it has the potential to make prediction at the exact same horizon as the original model.

%--------------------------------------------------
\section{The quasi-geostrophic model}
\label{sec:sec3-qg-model}
%--------------------------------------------------

In the following sections, we illustrate the potential of the DA--ML method to correct model error using twin experiments with a low-order model. For simplicity, the correction will be applied to the resolvent, as presented in Section~\ref{sssec:sec2-correcting-resolvent}. The model is a two-layer quasi-geostrophic channel model based on the equations of \citet{fandry-1984}. In the ECMWF OOPS-IFS, it captures the main dynamical processes which characterise extratropical geophysical flows while being of sufficient size and complexity to provide a convincing demonstration of the numerical methods tested in this article \citep{fisher-2017}.

%--------------------------------------------------
\subsection{Brief model description}
%--------------------------------------------------

\begin{figure}[tb]
    \centering
    \includegraphics[scale=\scale]{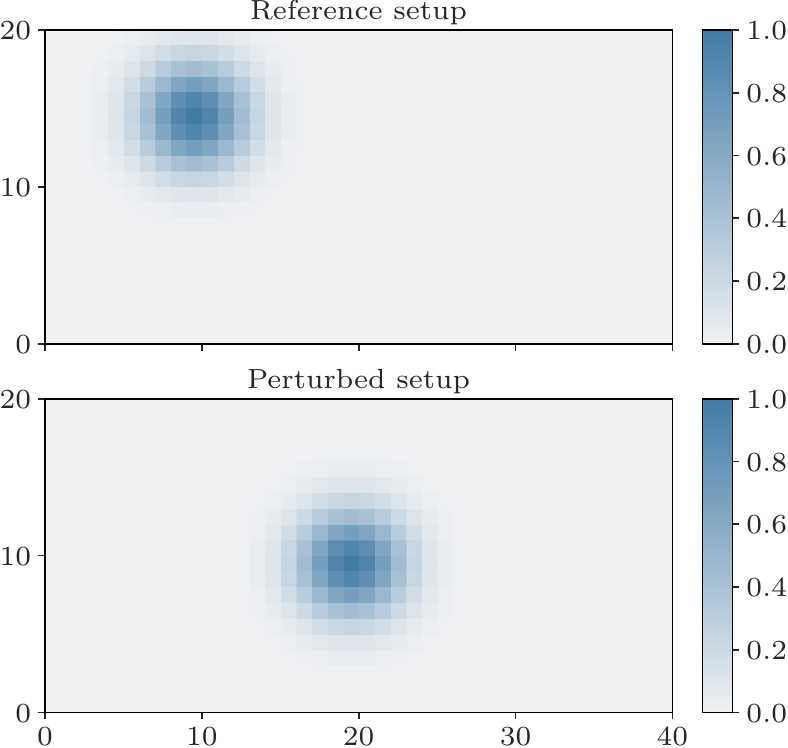}
    \caption{Non-dimensionalised orography for the reference setup (top panel) and the perturbed setup (bottom panel).}
    \label{fig:sec3-orography}
\end{figure}

The quasi-geostrophic (QG) model expresses the conservation of (non-dimensional) potential vorticity $q$ for two layers of constant potential temperature in the $x-y$ plane \citep{pedlosky-1987}:
\begin{equation}
    \frac{\mathrm{d}q_{1}}{\mathrm{d}t} = \frac{\mathrm{d}q_{2}}{\mathrm{d}t} = 0,
\end{equation}
where the subscripts $1$ and $2$ refer to the upper and lower layer, respectively. The potential vorticity $q$ is related to the stream function $\psi$ through
\begin{align}
    q_{1} &= \Delta \psi_{1} - F_{1} \left( \psi_{1} - \psi_{2} \right) + \beta y,\\
    q_{2} &= \Delta \psi_{2} - F_{2} \left( \psi_{2} - \psi_{1} \right) + \beta y + R_{\mathsf{s}}\left(x, y\right),
\end{align}
where $\beta$ is the (non-dimensionalised) northward derivative of the Coriolis parameter, $R_{\mathsf{s}}$ is the orography term, and $F_{1}$ and $F_{2}$ are two coupling parameters. The domain is periodic in the $x$ direction, and fixed boundary conditions are used for the potential vorticity $q$ in the $y$ direction. We use an horizontal discretisation of $\num{40}$ grid points in the $x$ direction and $\num{20}$ in the $y$ direction, and the orography is characterised by a Gaussian hill, as shown in the top panel of Figure~\ref{fig:sec3-orography}. For the numerical experiments with this model (forecast or DA), the control variable is chosen to be the stream function $\psi$ on both layers. Hence the state space has dimension $\num{40}\times\num{20}\times\num{2}=\num{1600}$. A complete model description, including all the parameter values, can be found in the work of \citet{laloyaux-2020}. Hereafter, this setup is called the \emph{reference setup}, as opposed to the \emph{perturbed setup} introduced in Section~\ref{ssec:sec3-perturbed-qg-model}.

%--------------------------------------------------
\subsection{Dynamical behaviour}
\label{ssec:sec3-dyn-behaviour}
%--------------------------------------------------

The integration scheme for the QG model - first-order, upstream, semi-Lagrangian for the potential vorticity $q$ - as well as its implementation \citep[see][]{laloyaux-2020} have been chosen for speed, stability, and convenience rather than for accuracy or conservation. Therefore, we first have to check the behaviour of the model over long integrations, which are needed to form the training database. 

\begin{figure}[tb]
    \centering
    \includegraphics[scale=\scale]{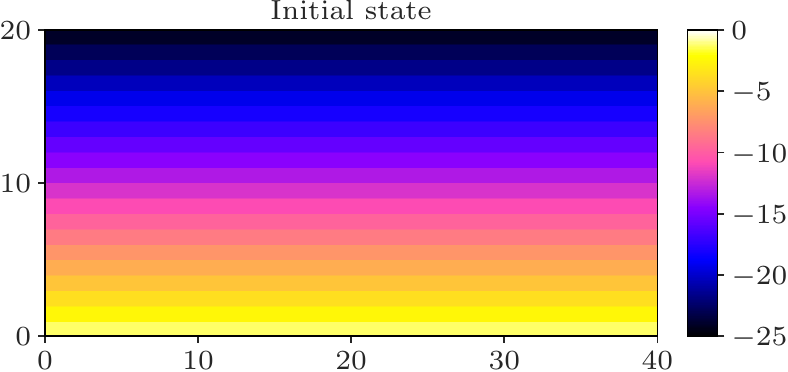}
    \caption{Initial state (stream function $\psi$) of the QG model in the reference setup. Only the bottom layer is shown, the stream function for the top layer following the exact same pattern.}
    \label{fig:sec3-initial-state}
\end{figure}

\begin{figure}[tb]
    \centering
    \includegraphics[scale=\scale]{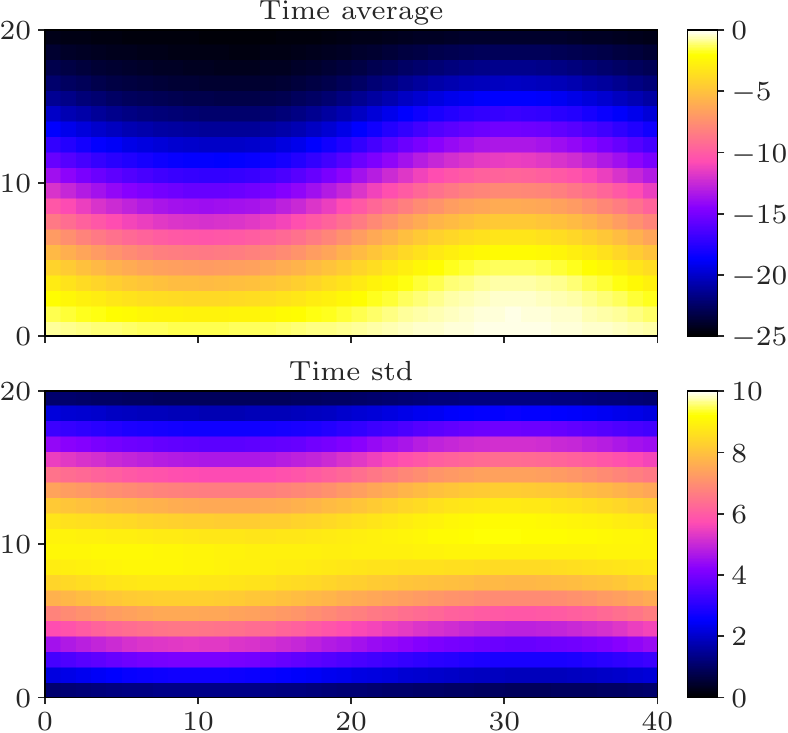}
    \caption{Climatology of the QG model in the reference setup. The top panel shows the time average and the bottom panel the time standard deviation. Only the bottom layer is shown, the climatology for the top layer following similar patterns.}
    \label{fig:sec3-climatology}
\end{figure}

For the reference setup, the initial state is shown in Figure~\ref{fig:sec3-initial-state}. After a relaxation period of several days, the evolution of $\psi$ is characterised by a wave, slowly moving towards the west, with a mean period around \SI{16}{\day}. The model climatology (time average and standard deviation) is shown in Figure~\ref{fig:sec3-climatology} and is closely related to the orography. The model variability (spatial average of the time standard deviation) is about $\num{4.95}$. We have also checked that an ensemble of forecasts yields the exact same climatology, which confirms the ergodicity of the model. The model is chaotic, with a doubling time of errors around \SI{250}{\hour}. For comparison, the doubling time of errors in the Integrated Forecasting System (IFS) developed at ECMWF is around \SI{2}{\day} (M. Bonavita, pers. comm.). Furthermore, the model is slightly less nonlinear than the IFS, having a longer duration of the linear regime before nonlinearities dominate the evolution \citep{gilmour-2001, fisher-2017}.

As one could expect, altering some model parameters such as the integration time step or the layer depths can yield a different dynamical behaviour, with a different  value for the doubling time of errors and a different climatology. This can explain the mismatch between the doubling time of errors reported here and the one reported by \citet{fisher-2017} for the same model. Surprisingly, starting from another initial condition can also yield a different dynamical behaviour. This suggests the existence of several distinct quasi-stationary regimes around the model attractor. In the numerical experiments, we have carefully checked that all the generated trajectories yield the same climatology described by Figure~{\ref{fig:sec3-climatology}}.

%--------------------------------------------------
\subsection{The perturbed QG model}
\label{ssec:sec3-perturbed-qg-model}
%--------------------------------------------------

For the numerical experiments shown in this article, the true dynamical model is set as the QG model in the reference setup. The original model is defined by adding some model error on top of this true model. Two options are possible for the model error.

In the weak-constraint 4D-Var test series by \citet{laloyaux-2020}, model error is introduced by an explicit forcing term added to the resolvent of the true QG model. The forcing term follows a zero-mean Gaussian distribution with given covariance matrix, and it is assumed constant in time. On the one hand, this approach enables an accurate characterisation of the model error and its statistical properties. On the other hand, very long forecasts with a constant forcing sometimes yield divergence. 

The other option is to introduce model error in the parameters. In the QG model, this can be implemented by changing the layer depths, the integration time step, or the orography. The idea is to mimic errors due to deficiencies in the physical parametrisation, limited spatial resolution, or erroneous time integration \citep{fisher-2017}. The model error here is implicit, and therefore harder to characterise, but we did not find any instability in the integration with this approach. This is why we choose to use it in the numerical experiments.

We introduce a perturbed setup for the QG model as a variant of the reference setup with four modifications. Both layer depths are changed, the integration time step is doubled, and the Gaussian hill (in the orography term) is moved to the centre of the domain. The reference and perturbed values are reported in Table~\ref{tab:sec3-parametrisation-ref-vs-perturbed} and Figure~\ref{fig:sec3-orography}.

\begin{table}[tb]
    \centering
    \caption{Set of parameters for the reference setup (middle row) and the perturbed setup (right row).}
    \label{tab:sec3-parametrisation-ref-vs-perturbed}
    \begin{tabular}{lrr}
    \toprule
    Parameter & Ref. setup & Pert. setup \\
    \midrule
    Top layer depth & \SI{6000}{\meter} & \SI{5750}{\meter} \\
    Bottom layer depth & \SI{4000}{\meter} & \SI{4250}{\meter} \\
    Integration time step & \SI{10}{\min} & \SI{20}{\min} \\
    \bottomrule 
    \end{tabular}
\end{table}

\begin{figure}[tb]
    \centering
    \includegraphics[scale=\scale]{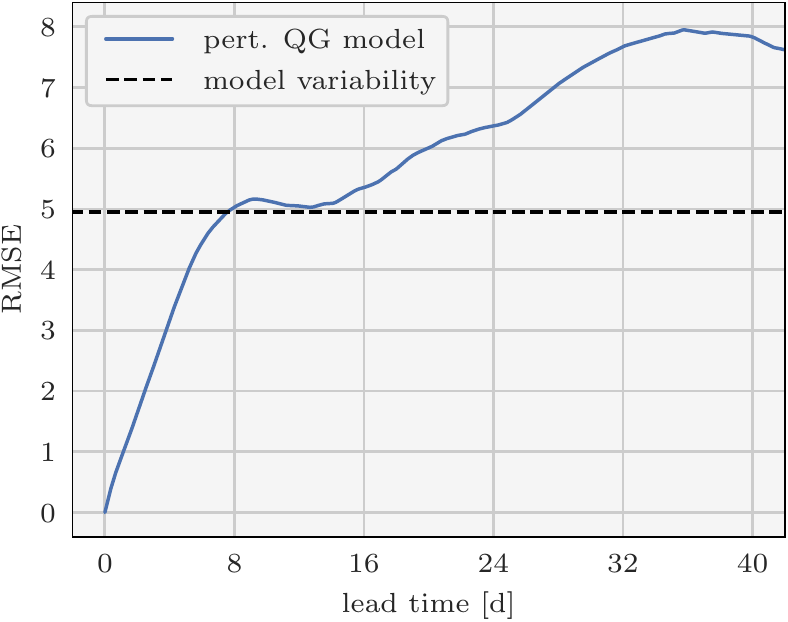}
    \caption{Forecast skill of the original (perturbed) model (continuous blue line) as a function of the lead time in days, computed using an ensemble of $N_{\mathsf{e}}=\num{100}$ initial conditions. For comparison, the model variability is shown with an horizontal, dashed black line.}
    \label{fig:sec3-forecast-skill}
\end{figure}

\begin{figure}[tb]
    \centering
    \includegraphics[scale=\scale]{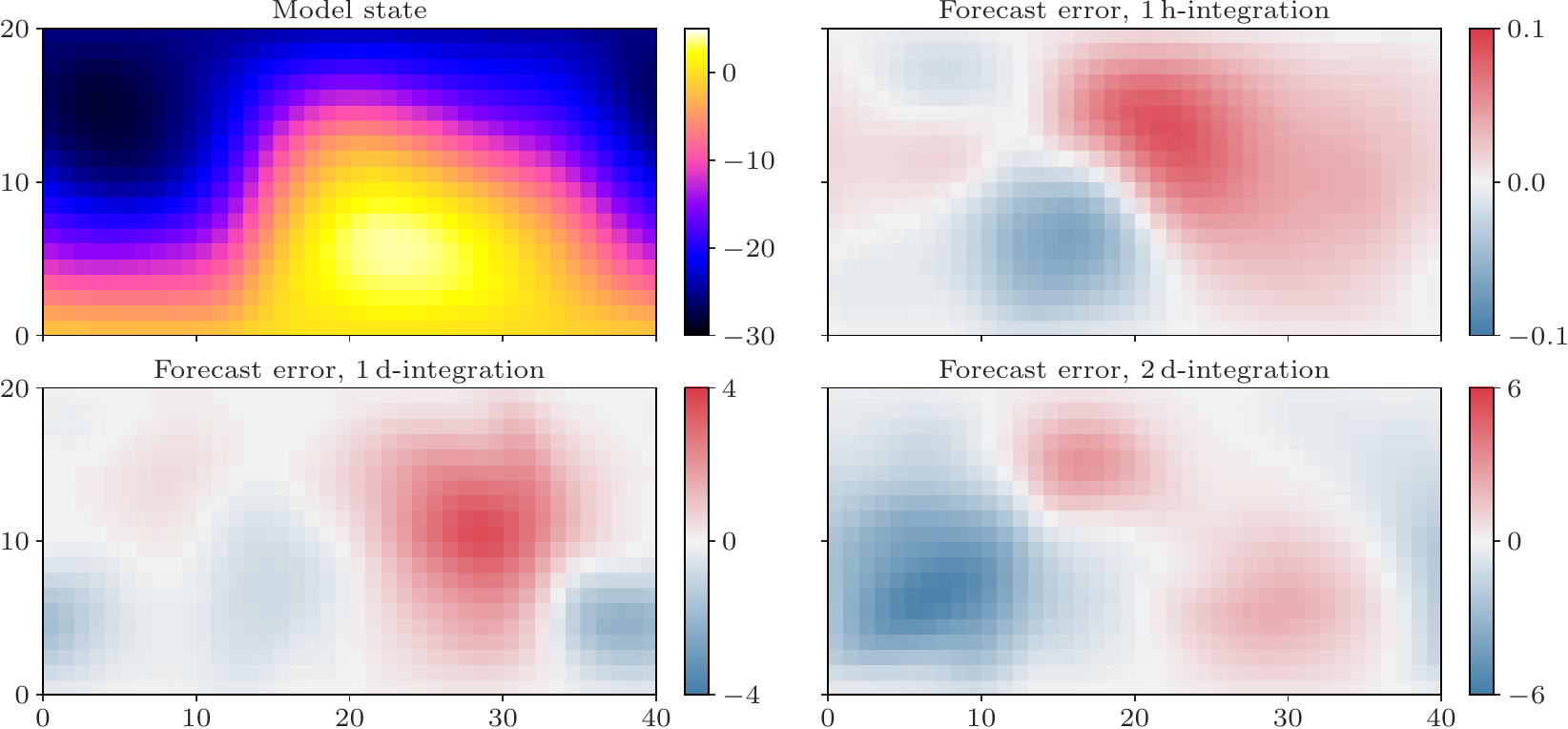}
    \caption{Snapshot of the QG model. The top left panel shows a typical model state in the reference setup. Starting from this model state, the top right panel shows the forecast error for a $\SI{1}{\hour}$-integration with the perturbed setup instead of the reference setup. The bottom panels show the same quantity for a $\SI{1}{\day}$-integration (bottom left) and for a $\SI{2}{\day}$-integration (bottom right). In all cases, only the bottom layer is shown.}
    \label{fig:sec3-snap}
\end{figure}

To summarise, the true model is the QG model in the reference setup, and the original model is the QG model in the perturbed setup. Figure~\ref{fig:sec3-forecast-skill} shows the forecast skill (FS) of the original model, defined as
\begin{equation}
    \label{eq:sec3-def-fs}
    \mathrm{FS}_{k} \triangleq \frac{1}{N_{\mathsf{e}}}\sum_{j=1}^{N_{\mathsf{e}}} \mathrm{RMSE}\left[\mathcal{M}^{\mathsf{t}}_{0\to k}\left(\mathbf{x}^{\mathsf{i}}_{j}\right),\mathcal{M}^{\mathsf{o}}_{0\to k}\left(\mathbf{x}^{\mathsf{i}}_{j}\right)\right],
\end{equation}
where $\mathcal{M}^{\mathsf{t}}_{0\to k}$ and $\mathcal{M}^{\mathsf{o}}_{0\to k}$ are the resolvent of the true and original QG model from $t_{0}$ to $t_{k}$, respectively, and $\mathbf{x}^{\mathsf{i}}_{1:N_{\mathsf{e}}}$ is a set of $N_{\mathsf{e}}$ initial conditions representative of the true model climatology\footnote{The QG model being autonomous, the actual starting date of these initial conditions does not matter and can be $t_{0}$.}. Finally for visualisation purposes, Figure~{\ref{fig:sec3-snap}} illustrates a snapshot of the QG model and the model error associated to the perturbed setup.

%--------------------------------------------------
\section{The original data assimilation step}
\label{sec:sec4-original-da-step}
%--------------------------------------------------

The DA--ML method starts with a DA step in which the model is the original model. In this section, we present the implementation and the results for this original DA step.

%--------------------------------------------------
\subsection{The observation database}
\label{ssec:sec4-observation-database}
%--------------------------------------------------

Starting from a set $\mathbf{x}^{\mathsf{i}}_{1:N_{\mathsf{e}}}$ of initial conditions representative of the true model climatology, we compute $N_{\mathsf{e}}$ trajectories of the true model. Using this catalogue of trajectories, an observation database is created as follows. Observations are available every $\Delta t=\SI{2}{\hour}$, starting at 01:00 every day. The observation operator $\mathcal{H}_{k}$ consists in a bilinear interpolation of the stream function $\psi$ at $N_{\mathsf{y}}=\num{50}$ random locations. The observation noise follows a zero-mean Gaussian distribution with covariance matrix $\mathbf{R}=\num{0.1}\mathbf{I}$ (about $\SI{2}{\percent}$ of the model variability). Hence, the system is partially observed, in a way which mimics the homogeneous coverage provided by satellite soundings. Overall, the observation frequency is very high compared to the doubling time or errors (around $\SI{250}{\hour}$ as mentioned in {\ref{ssec:sec3-dyn-behaviour}}) but this is compensated by the high sparsity of observation. In the end, the data assimilation problem is reasonably complex, especially taken into account the relatively high model error.

For the numerical experiments shown in the following sections, the database is made of $N_{\mathsf{e}}=\num{18}$ trajectories. In the ML step, the first trajectory is used for training, the second is used for validation, and the $\num{16}$ remaining are used for testing. Differently to most ML studies, much more data is used for testing in order to get reliable testing metrics.  

%--------------------------------------------------
\subsection{Data assimilation setup}
%--------------------------------------------------

The observations are assimilated using the (strong-constraint) 4D-Var algorithm, with consecutive data assimilation windows (DAWs) of $\Delta T=\SI{1}{\day}$ starting at 00:00 each. Each 4D-Var problem consists in minimising the cost function
\begin{equation}
    \label{eq:sec4-def-cost-4dvar}
    \mathcal{J}\left(\mathbf{x}^{\mathsf{i}}\right) 
    = \frac{1}{2} \left\| \mathbf{x}^{\mathsf{i}} - \mathbf{x}^{\mathsf{b}} \right\|_{\mathbf{B}^{-1}}
    + \frac{1}{2} \sum_{k=1}^{12} \left\| \mathbf{y}_{k} - \mathcal{H}_{k} \circ \mathcal{M}^{\mathsf{o}}_{0\to k}\left(\mathbf{x}^{\mathsf{i}}\right) \right\|_{\mathbf{R}^{-1}},
\end{equation}
where $\mathbf{x}^{\mathsf{i}}$ is the unknown state at the start of the window, $\mathbf{x}^{\mathsf{b}}$ is the background state, $\mathcal{M}^{\mathsf{o}}_{0\to k}$ is the resolvent of the original model from $t_{0}$ (start of the window) to $t_{k}$ (time of the $k$-th batch of observations), $\mathbf{R}$ is the observation error covariance matrix as defined in Section~\ref{ssec:sec4-observation-database}, and $\mathbf{B}$ is the background error covariance matrix.

Since the true trajectories are known, the main performance criterion is the root mean squared error (RMSE) between the analysis and the truth at the start of the window, simply called the analysis RMSE. In this cycled context, the main degree of freedom for the DA is the background error covariance matrix $\mathbf{B}$. We choose to use $\mathbf{B} = b^{2}\mathbf{C}$, where $\mathbf{C}$ is a correlation matrix with specified horizontal correlation length and vertical correlation, and $b$ is the standard deviation. The values of these parameters are optimally tuned to yield the lowest time-averaged analysis RMSE and reported in Table~\ref{tab:sec4-parametrisation-B}. Overall, $\mathbf{B}$ has short-range correlation, which is necessary to correct short-range patterns in the background error \citep{laloyaux-2020}.

\begin{table}[tb]
    \centering
    \caption{Optimal parametrisation of the background error covariance matrix $\mathbf{B}$. The horizontal correlation length is given in fraction of the horizontal domain length.}
    \label{tab:sec4-parametrisation-B}
    \begin{tabular}{lr}
    \toprule
    Parameter & Value \\
    \midrule
    Hor. corr. len. & $\num{0.6}$ \\
    Vert. corr. & $\num{0.2}$ \\
    Std. dev. $b$ & $\num{0.08}$ \\
    \bottomrule 
    \end{tabular}
\end{table}

%--------------------------------------------------
\subsection{Data assimilation results}
\label{ssec:sec4-da-results}
%--------------------------------------------------

This method is applied to all $N_{\mathrm{e}}=\num{18}$ trajectories. A total of $\num{1032}$ consecutive cycles are performed without numerical failure for each trajectory. The analysis RMSE stabilises after about $\num{5}$ cycles. Hence the first $\num{8}$ cycles are dropped as spin-up cycles, and the time-averaged analysis RMSE, averaged over the $N_{\mathrm{e}}=\num{18}$ trajectories, is $\num{0.24}$ (about $\SI{5}{\percent}$ of the model variability).

%--------------------------------------------------
\section{The machine learning step}
\label{sec:sec5-ml-step}
%--------------------------------------------------

The result of the DA is a database of analysis trajectories, from which we can now perform the first ML step. In this section, we present the numerical choices and the results for this ML step.

%--------------------------------------------------
\subsection{Database preparation}
\label{ssec:sec5-database}
%--------------------------------------------------

Consider one of the $N_{\mathsf{e}}$ analysis trajectories, and let $\mathbf{x}^{\mathsf{a}}_{k}$ be the analysis state at the start of the window for the $k$-th cycle. Note that contrary to the notation adopted in Section~\ref{sec:sec2-methodology}, the time index $k$ refers here to the cycle number and not to the observation time (each cycle corresponds to one DAW, which gathers $\num{12}$ observation batches). Let $\mathbf{x}^{\mathsf{t}}_{k}$ be the corresponding true state. The resolvent of the true model for a $1$-window ($\Delta T=\SI{1}{\day}$) integration, $\mathcal{M}^{\mathsf{t}}$, satisfies
\begin{equation}
    \mathbf{x}^{\mathsf{t}}_{k+1} = \mathcal{M}^{\mathsf{t}}\left(\mathbf{x}^{\mathsf{t}}_{k}\right).
\end{equation}
Following the rationale described in Section~\ref{sssec:sec2-correcting-resolvent}, $\mathcal{M}^{\mathsf{t}}$ is approximated by $\mathcal{M}^{\mathsf{o}}+\mathcal{M}^{\mathsf{ml}}$, where $\mathcal{M}^{\mathsf{o}}$ is the resolvent of the original model for a $1$-window integration, and $\mathcal{M}^{\mathsf{ml}}$ is the ML model. The true (exact) database to learn the model error is therefore
\begin{equation}
    \mathcal{D}^{\mathsf{t}}\left(N_{\mathsf{t}}\right) \triangleq \Big\{\Big(\mathbf{x}^{\mathsf{t}}_{k},\; \mathbf{x}^{\mathsf{t}}_{k+1}-\mathcal{M}^{\mathsf{o}}\left(\mathbf{x}^{\mathsf{t}}_{k}\right)\Big),\; k=1\ldots N_{\mathsf{t}}\Big\},
\end{equation}
where $\mathrm{d}\mathbf{x}^{\mathsf{t}}_{k}\triangleq\mathbf{x}^{\mathsf{t}}_{k+1}-\mathcal{M}^{\mathsf{o}}\left(\mathbf{x}^{\mathsf{t}}_{k}\right)$ is the 1-window \emph{model error} and $N_{\mathsf{t}}$ is the number of samples in the database. However, the truth is unknown and its best available approximation is the analysis. This is why the effective database for the ML step is
\begin{equation} \label{db1}
    \mathcal{D}^{\mathsf{a}}\left(N_{\mathsf{t}}\right) \triangleq \Big\{\Big(\mathbf{x}^{\mathsf{a}}_{k},\; \mathbf{x}^{\mathsf{a}}_{k+1}-\mathcal{M}^{\mathsf{o}}\left(\mathbf{x}^{\mathsf{a}}_{k}\right)\Big),\; k=1\ldots N_{\mathsf{t}}\Big\},
\end{equation}
where, in this cycled DA context, $\mathrm{d}\mathbf{x}^{\mathsf{a}}_{k}\triangleq\mathbf{x}^{\mathsf{a}}_{k+1}-\mathcal{M}^{\mathsf{o}}\left(\mathbf{x}^{\mathsf{a}}_{k}\right)$ is the \emph{analysis increment} computed in the data assimilation step. The effective database $\mathcal{D}^{\mathsf{a}}\left(N_{\mathsf{t}}\right)$ is a good approximation of the true database $\mathcal{D}^{\mathsf{t}}\left(N_{\mathsf{t}}\right)$ if the analysis approximates the true model state ($\mathbf{x}^{\mathsf{a}}_{k}\approx\mathbf{x}^{\mathsf{t}}_{k}$) and if the analysis increment approximates the true model error ($\mathrm{d}\mathbf{x}^{\mathsf{a}}_{k}\approx\mathrm{d}\mathbf{x}^{\mathsf{t}}_{k}$). Of course, if $\mathcal{M}^{\mathsf{o}}$ is continuous, which is a reasonable assumption, then $\mathbf{x}^{\mathsf{a}}_{k}\approx\mathbf{x}^{\mathsf{t}}_{k}$ implies $\mathrm{d}\mathbf{x}^{\mathsf{a}}_{k}\approx\mathrm{d}\mathbf{x}^{\mathsf{t}}_{k}$. However, this does not ensure that the relative accuracy of both approximations are of similar order. This point is discussed in details in Section~{\ref{ssec:sec5-changing-database-hyperparams}}.

The preprocessing is repeated $N_{\mathsf{e}}=\num{18}$ times, once for every analysis trajectory, which yields $\num{18}$ databases $\mathcal{D}^{\mathsf{a}}\left(N_{\mathsf{t}}\right)$. As previously stated, the first one is used for training, the second one is used for validation, and the other $\num{16}$ ones are used for testing. The default size of the database used for the training is $N_{\mathsf{t}}=\num{1024}$, which means that the database contains the analyses and increments of all $\num{1024}$ consecutive DAWs.

For completeness, we would like to mention the fact that using the analysis increments to estimate model error is an idea which has originally been suggested by \citet{leith-1978} and later by \citet{carrassi-2011} in the context of unresolved scales. It has already been applied, \textit{e.g.}, by \citet{mitchell-2015} with the increments of an ensemble Kalman filter. However, where their approach focus on the first two statistical moments of the increments, our approach is more ambitious as it aims at learning the entire relationship between the analysis and the analysis increment.

%--------------------------------------------------
\subsection{Machine learning models architecture}
\label{ssec:sec5-ml-template}
%--------------------------------------------------

We choose to build the ML model using NNs. The method needs to be scalable to high-dimension problems, which is why, for this small problem (both input and output space of the models have dimension $N_{\mathsf{x}}=\num{1600}$), the NN should be as simple and generic as possible, and use as few parameters as possible. Therefore, we consider sequential models composed by \emph{fully-connected} or \emph{dense} (DNN) layers. The DNN layer is the standard building block for NNs. As indicated by its name, it connects all output variables to all input variables. However, in many realistic models, the correlations decrease with the physical distance between variables at a fast rate, which explains the success of localisation methods for ensemble DA. For this reason, we also consider replacing some DNN layers by \emph{convolutional} (CNN) layers, in which only local operations are performed. 

\begin{figure}[tb]
    \centering
    \includegraphics[scale=\scale]{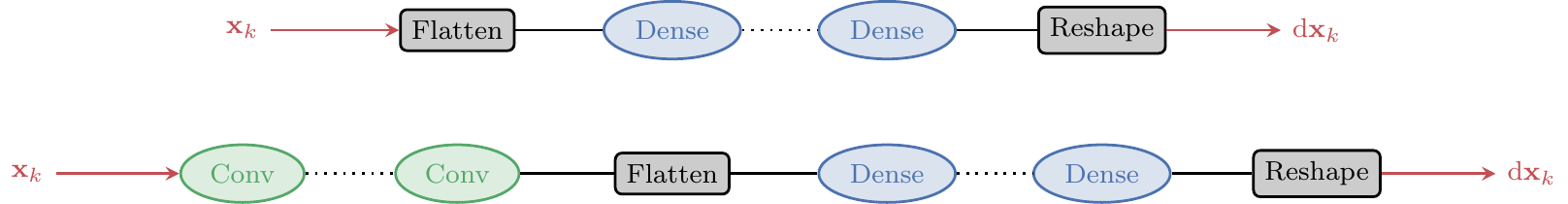}
    \caption{Template for the D models (top panel) and the CD models (bottom panel). The convolutional layers are two-dimensional, each model layer is treated as a filter, and the periodic boundary conditions are taken into account by adding enough padding in the input. Finally, flatten and reshape layers are added before and after the dense layers to make the interface between three-dimensional and one-dimensional fields.}
\label{fig:sec5-networks}
\end{figure}

In summary, two NN architectures are tested: sequential models with only DNN layers (D models) and sequential models with CNN layers followed by DNN layers (CD models). This is illustrated by Figure~\ref{fig:sec5-networks}. Beyond the layer types, the other NN parameters, including the activation functions, are discussed in Section~\ref{ssec:sec5-comparison-ml-models}. All NNs are implemented using TensorFlow 2.x \citep{abadi-2016}. They are trained with Adam, a variant of the stochastic gradient descent \citep{kingma-2017}, using the mean squared error (MSE) as loss function. The training consists of $\num{e3}$ epochs with an initial learning rate of $\num{e-3}$, followed by $\num{e3}$ epochs with an initial learning rate of $\num{e-4}$. In each case, the model with the lowest validation MSE is kept. Furthermore, the input and output are normalised in order to accelerate the convergence. This minimisation strategy is empirically adequate for all the experiments shown in this paper.

%--------------------------------------------------
\subsection{Training example}
\label{ssec:sec5-training-example}
%--------------------------------------------------

Let us consider the simplest example of a D model made of $1$ DNN layer with $4$ nodes and linear activation functions. This NN has a total of $\num{14404}$ parameters: $\num{1600}\times\num{4}+\num{4}=\num{6404}$ parameters for the connection between the input and the DNN layers, and $\num{4} \times\num{1600}+\num{1600}=\num{8000}$ for the connection between the DNN and output layers. Following the method described in Section~\ref{ssec:sec5-ml-template}, the NN is trained with $\mathcal{D}^{\mathsf{a}}\left(\num{128}\right)$, which implies that the size of the database is $N_{\mathsf{t}}=\num{128}$\footnote{The choice of $N_{\mathsf{t}}=\num{128}$ is intermediate and leaves the option to use either smaller or larger databases.}. After the training, the normalised test MSE, averaged over all $\num{16}$ test trajectories and corrected by the trajectory variance, is $\SI{17.43}{\percent}$, with a standard deviation over the $\num{16}$ trajectories of $\SI{0.77}{\percent}$.

\begin{figure*}[tb]
    \centering
    \includegraphics[scale=\scale]{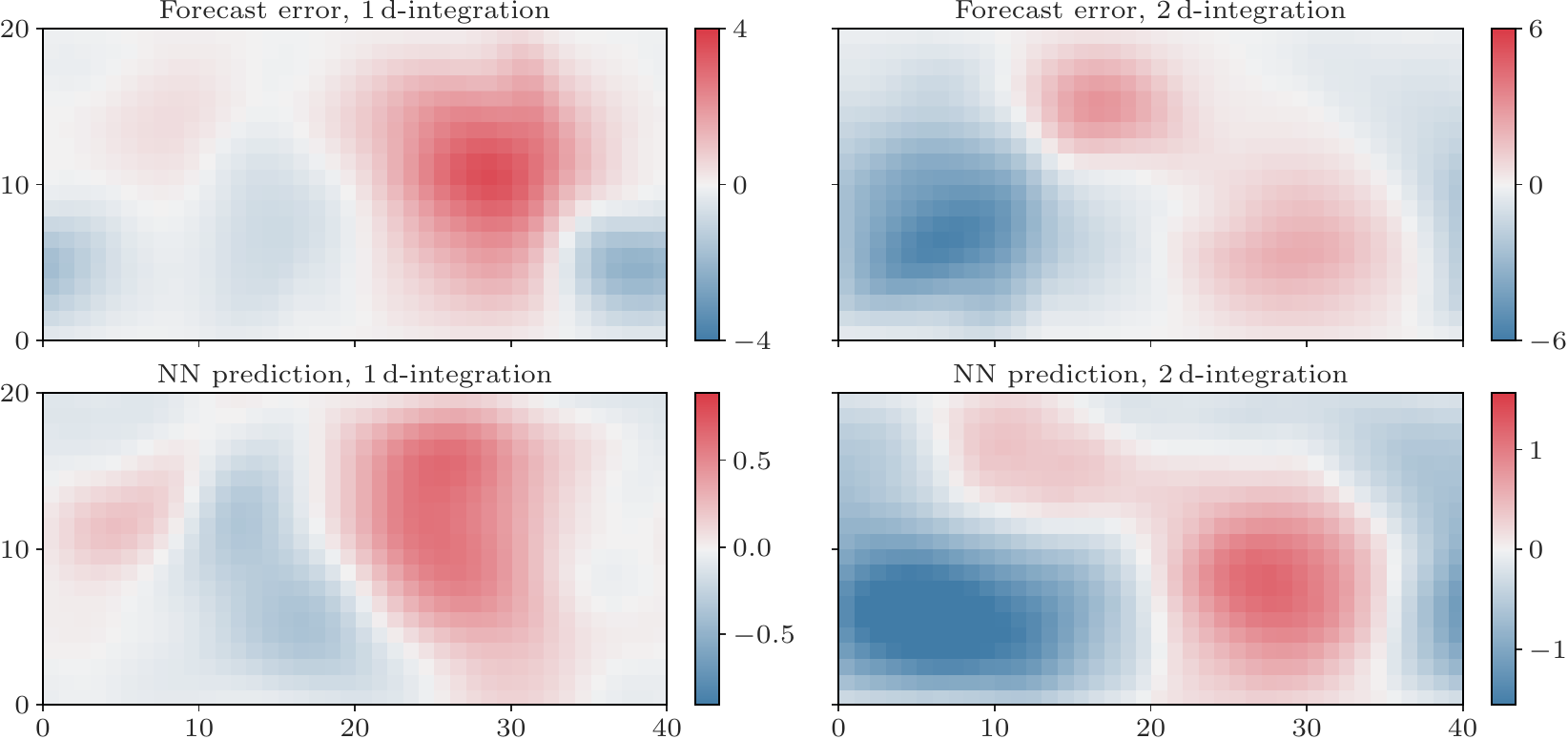}
    \caption{Example of model error prediction. The top left panel shows the forecast error for a $\SI{1}{\day}$-integration starting from the model state shown in Figure~\ref{fig:sec3-snap}. The bottom left panel shows the corresponding prediction with the NN described in Section~\ref{ssec:sec5-training-example}. The right panels show the same quantities for a $\SI{2}{\day}$-integration, \textit{i.e.} when the sampling period $\tau$ has been doubled as explained in Section~\ref{ssec:sec5-changing-database-hyperparams}. In all cases, only the bottom layer is shown.}
    \label{fig:sec5-snap}
\end{figure*}

However, the objective of the NN is not to predict the analysis increment but the model error. Therefore, keeping $\mathcal{D}^{\mathsf{a}}\left(\num{128}\right)$ for the training, we evaluate the NN performance using the corresponding true database $\mathcal{D}^{\mathsf{t}}\left(\num{128}\right)$. In this case, we obtain a normalised test MSE of $\SI{68.50}{\percent}$. An example of model error prediction with this NN is illustrated on the left panels of Figure~\ref{fig:sec5-snap}. For this prediction example, the scaling (spatial average and standard deviation) are reported in Table~\ref{tab:sec5-training-example-scaling}. Visually, the NN is able to predict the correct model error patterns but the scaling is incorrect.
There is a significant difference between the analysis increment predictions and the model error predictions, which is related to the accuracy of the approximation $\mathrm{d}\mathbf{x}^{\mathsf{t}}_{k}\approx\mathrm{d}\mathbf{x}^{\mathsf{a}}_{k}$. This point is further discussed in Section~\ref{ssec:sec5-changing-database-hyperparams}.

\begin{table}[tb]
    \centering
    \caption{Scaling (spatial average and standard deviation) of the model error prediction example shown in Figure~\ref{fig:sec5-snap}. The first line reports the values for the exact forecast error (top panels of Figure~\ref{fig:sec5-snap}), the middle line reports the values for the NN prediction (bottom panels of Figure~\ref{fig:sec5-snap}), and the bottom line reports the relative error.}
    \label{tab:sec5-training-example-scaling}
    \begin{tabular}{lrrrr}
    \toprule
    Field & Mean & Std & Mean & Std \\
    & \multicolumn{2}{c}{$\SI{1}{\day}$-integration} & \multicolumn{2}{c}{$\SI{2}{\day}$-integration} \\
    \midrule
    Exact forecast error & $\num{0.31}$ & $\num{0.93}$ & $\num{-0.57}$ & $\num{1.46}$ \\
    NN prediction & $\num{0.05}$ & $\num{0.26}$ & $\num{-0.24}$ & $\num{0.63}$ \\
    Relative error & $\SI{84}{\percent}$ & $\SI{72}{\percent}$ & $\SI{58}{\percent}$ & $\SI{56}{\percent}$\\
    \bottomrule 
    \end{tabular}
\end{table}

\begin{figure}[tb]
    \centering
    \includegraphics[scale=\scale]{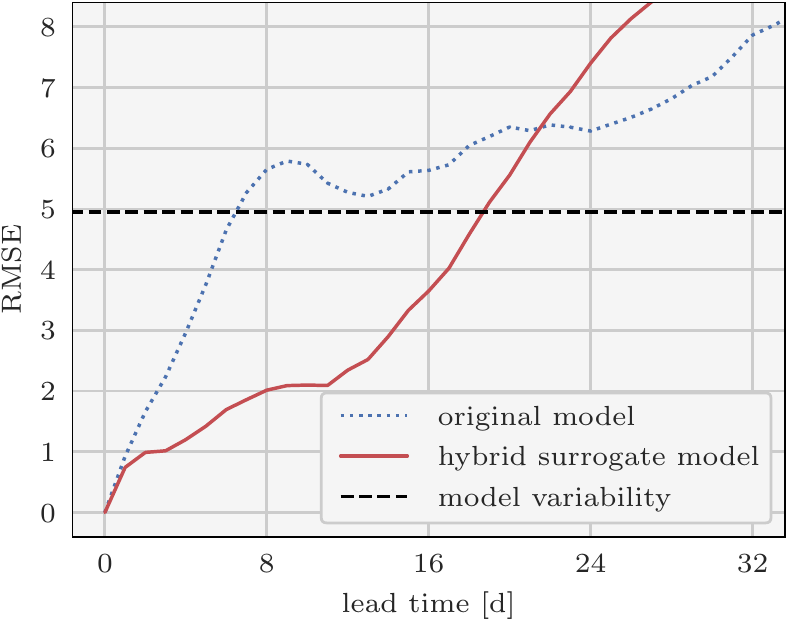}
    \caption{Forecast skill of the original model (dotted blue line) and of the hybrid surrogate model (continuous red line) as a function of the lead time in days, computed using an ensemble of $N_{\mathrm{e}}=16$ initial conditions. For comparison, the model variability is shown with an horizontal, dashed black line.}
    \label{fig:sec5-forecast-skill}
\end{figure}

The test MSE is a good metric to evaluate the ability of the NN to correct the original model in making a short-range forecast of $\SI{1}{\day}$. As a complement, we want to evaluate the accuracy of the mid- to long-range forecasts. To do this, we compute the FS of the hybrid surrogate model, defined by Equation~\eqref{eq:sec3-def-fs} in which $\mathcal{M}^{\mathsf{o}}$, the resolvent of the original model, is to be replaced by $\mathcal{M}^{\mathsf{o}}+\mathcal{M}^{\mathsf{ml}}$, the resolvent of the hybrid surrogate model. This is shown in Figure~\ref{fig:sec5-forecast-skill}. Overall, the NN is indeed able to correct the original model, in such a way that the hybrid surrogate model yields better forecasts up to about $\SI{16}{\day}$. It is also remarkable that the correction seems to be more effective for mid-range (\SIrange{3}{10}{\day}) than for short-range (\SIrange{1}{3}{\day}) forecasts. A further study is required to understand this phenomenon. In particular, we need to check whether it is related to the use of analysis increments as training data, or to the specific model and setup in any way.

Finally, we conclude this section by reporting the execution time of this ML step. The computation has been performed using a Bi-Xeon Gold 6154 (18 cores) without a GPU card. The entire training phase took $\SI{31}{\second}$ (wall-clock time) for a total of $\num{2000}$ epochs. We have also performed the exact same training phase using a Xeon W-2133 (only 6 cores) with a GPU card (in this case an Nvidia Quadro RTX5000), but the execution time did not improve. Indeed, since the training data is overall small, the overhead related to the use of a GPU card is higher than the acceleration in the linear algebra. For comparison, the entire original DA step ($\num{1032}$ windows) for one trajectory took $\SI{9233}{\second}$ using only one thread of the same Bi-Xeon Gold 6154.

%--------------------------------------------------
\subsection{Improving the quality of the training database}
\label{ssec:sec5-changing-database-hyperparams}
%--------------------------------------------------

We have shown in Section~\ref{ssec:sec5-training-example} that the analysis increment predictions by the NN are much more accurate than the model error predictions. From a physical perspective, the analysis increments are used as a proxy for the model error. However, the analysis increments computed by any DA method are affected by additional sources of errors beyond model errors \citep{dee-2005}. The variability of the analysis increments generally depends on the amount of information extracted from the observations, but this relationship is not straightforward. Small analysis increments can be the sign of a very good forecast model, but they can be due simply to a lack of observations \citep{dee-2011, crawford-2020}. Increments can also be affected by the approximations in the observation operators or by the DA method itself.

From a mathematical perspective, the difference between analysis increment and the model error predictions is related to the nature and accuracy of the training database $\mathcal{D}^{\mathsf{a}}\left(N_{\mathsf{t}}\right)$. Indeed, the ML step relies on the accuracy of the analysis $\mathbf{x}^{\mathsf{a}}_{k}$ used in the input $\mathbf{x}^{\mathsf{a}}_{k}$ and the output $\mathrm{d}\mathbf{x}^{\mathsf{a}}_{k}=\mathbf{x}^{\mathsf{a}}_{k+1}-\mathcal{M}^{\mathsf{o}}\left(\mathbf{x}^{\mathsf{a}}_{k}\right)$ to approximate respectively the true state $\mathbf{x}^{\mathsf{t}}_{k}$ and the exact model error $\mathrm{d}\mathbf{x}^{\mathsf{t}}_{k}=\mathbf{x}^{\mathsf{t}}_{k+1}-\mathcal{M}^{\mathsf{o}}\left(\mathbf{x}^{\mathsf{t}}_{k}\right)$. The original model is integrated here over a single DAW of $\SI{1}{\day}$, in such a way that $\mathrm{d}\mathbf{x}^{\mathsf{a}}_{k}$ is the analysis increment computed in the DA step. This makes the training database particularly sensitive to the quality of the analysis as the model evolves over a rather short period of time. Indeed, we have seen in Section~\ref{ssec:sec4-da-results} that the average RMS norm of the analysis error $\mathbf{x}^{\mathsf{a}}_{k}-\mathbf{x}^{\mathsf{t}}_{k}$ is $\num{0.24}$ (analysis RMSE), while Figure~{\ref{fig:sec3-forecast-skill}} shows that the average RMS norm of the true model error $\mathrm{d}\mathbf{x}^{\mathsf{t}}_{k}$ for a $\SI{1}{\day}$-forecast  is around $\num{1}$ (FS at $\SI{1}{\day}$). In order to reduce this sensitivity, we introduce a new family of databases where the model is integrated over several DAWs. Such databases are expected to contain more of the model error dynamics by capturing the model drift towards its preferred climatology in the medium-range, while being less sensitive to the quality of the analysis. Indeed, as the model integration gets longer, the model error increases (exceeding the model variability of $\num{4.95}$ after $\SI{8}{\day}$ as can be seen in Figure~\ref{fig:sec3-forecast-skill}) and becomes much more significant that the analysis RMSE ($\num{0.24}$).
%It will also be less sensitive to the quality of the analysis and to spin-up effects happening at any model initialization that are possibly not linked to model biases.

\begin{figure*}[tb]
    \centering
    \includegraphics[scale=\scale]{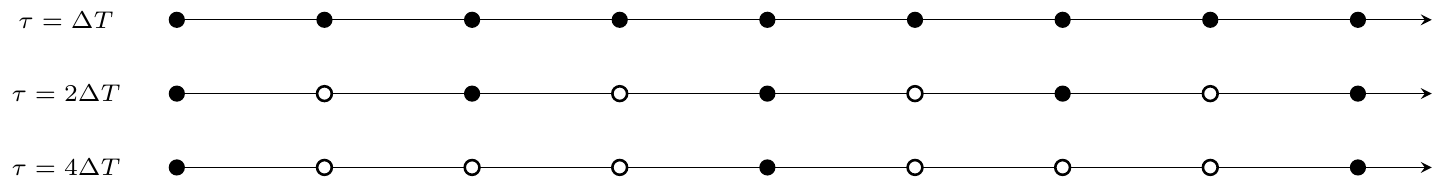}
    \caption{Illustration of the construction of the ML database $\mathcal{D}^{\mathsf{a}}\left(\tau, N_{\mathsf{t}}\right)$. Each circle represents one analysis sample $\mathbf{x}^{\mathsf{a}}_{k}$, corresponding to one DAW. Filled circles are kept in the ML database, while empty circles are discarded. For $\tau=\Delta T$ (top line), all analysis samples are kept and $N_{\mathsf{t}}=9$. For $\tau=2\Delta T$ (middle line), we keep one analysis sample out of two and $N_{\mathsf{t}}=5$. Finally for $\tau=4\Delta T$ (bottom line), we keep one analysis sample out four and $N_{\mathsf{t}}=3$.}
    \label{fig:sec5-sampling}
\end{figure*}

Effectively, the new exact and effective ML databases are defined by
\begin{align}
    \mathcal{D}^{\mathsf{t}}\left(\tau, N_{\mathsf{t}}\right) = \mathcal{D}^{\mathsf{t}}\left(n\Delta T, N_{\mathsf{t}}\right)&\triangleq \Big\{\Big(\mathbf{x}^{\mathsf{t}}_{nk},\; \mathbf{x}^{\mathsf{t}}_{n(k+1)}-\mathcal{M}^{\mathsf{o}}\left(\mathbf{x}^{\mathsf{t}}_{nk}\right)\Big),\; k=1\ldots N_{\mathsf{t}}\Big\},\\
    \mathcal{D}^{\mathsf{a}}\left(\tau, N_{\mathsf{t}}\right) = \mathcal{D}^{\mathsf{a}}\left(n\Delta T, N_{\mathsf{t}}\right) &\triangleq \Big\{\Big(\mathbf{x}^{\mathsf{a}}_{nk},\; \mathbf{x}^{\mathsf{a}}_{n(k+1)}-\mathcal{M}^{\mathsf{o}}\left(\mathbf{x}^{\mathsf{a}}_{nk}\right)\Big),\; k=1\ldots N_{\mathsf{t}}\Big\},
\end{align}
where $\tau=n\Delta T$ is the length of the individual model integrations used to capture the model error and $N_{\mathsf{t}}$ is the number of samples in the database. Several examples are illustrated by Figure~\ref{fig:sec5-sampling}. From now on, $\tau$ is referred to as the \emph{sampling period} of the ML step. The true database $\mathcal{D}^{\mathsf{t}}\left(\tau, N_{\mathsf{t}}\right)$ now captures the model error developing over several days. The effective database $\mathcal{D}^{\mathsf{a}}\left(\tau, N_{\mathsf{t}}\right)$ is based on model trajectories initialised from analyses, but it does not contain analysis increments from the DA step any more.

\begin{figure*}[tb]
    \centering
    \includegraphics[scale=\scale]{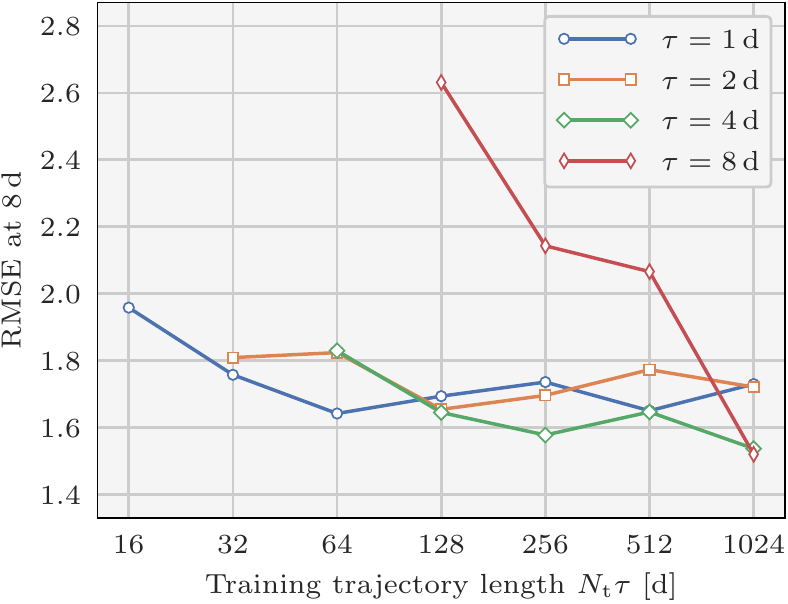}
    \caption{Evolution of the FS at $\SI{8}{\day}$ as a function of the total length of the training trajectory $N_{\mathsf{t}}\tau$ in number of days for several values of the sampling period $\tau$. For each experiment, the FS is computed using an ensemble of $N_{\mathrm{e}}=\num{16}$ initial conditions, and we select the NN which yields the lowest FS at $\SI{8}{\day}$ over all $\num{24}$ NNs.}
    \label{fig:sec5-summary-8}
\end{figure*}

Figure~\ref{fig:sec5-snap} and Table~\ref{tab:sec5-training-example-scaling} illustrate the effect of doubling the sampling period $\tau$ in a model error prediction with the same NN. In particular, the relative errors on the scaling are significantly reduced. In this section, we further illustrate the influence of both the sampling period $\tau$ and the database length $N_{\mathsf{t}}$ on the FS at $\SI{8}{\day}$. A forecast of $\SI{8}{\day}$ corresponds to a typical mid-range forecast, at the same time long enough to make the original model diverge from the true model (FS higher than the model variability as can be seen in Figure~\ref{fig:sec3-forecast-skill}) and short enough to expect reasonably good FSs with the hybrid surrogate models. To do this, we train an ensemble of $\num{24}$ NNs with $\mathcal{D}^{\mathsf{a}}\left(\tau, N_{\mathsf{t}}\right)$ for several values of $\tau$ and $N_{\mathsf{t}}$. The $\num{24}$ NNs consist of all possible D and CD models with (i) $\num{1}$ or $\num{4}$ layers of each type, (ii) $\num{4}$, $\num{8}$, or $\num{16}$ nodes/filters per layer, and (iii) linear or relu activation\footnote{Relu activation means that each layer, except for the very last one, uses $x\mapsto \mathrm{max}(0,x)$ as activation function.}. For each training configuration, we select the NN which yields the best results over all $\num{24}$ NNs. Actual comparisons between the different NNs are made in Section~\ref{ssec:sec5-comparison-ml-models}. Figure~\ref{fig:sec5-summary-8} shows the evolution of the FS at $\SI{8}{\day}$ as a function of the total length of the training trajectory $N_{\mathsf{t}}\tau$. The results depicted by Figure~\ref{fig:sec5-summary-8} could seem surprising, because one would expect two tendencies: (i) at fixed sampling period $\tau$, the FS improves as the training trajectory gets longer because there is more statistical information in longer trajectories and (ii) globally the FS improves as the sampling period $\tau$ gets higher because it is intuitively easier to correct short-term model errors (especially taken into account that the dynamical model is chaotic). By contrast, Figure~\ref{fig:sec5-summary-8} shows that, except for the extreme case $\tau=\SI{8}{\day}$, the FS at $\SI{8}{\day}$ is approximately the same for a wide range of values for $N_{\mathsf{t}}$ and $\tau$. We draw the following conclusions. First, short trajectories are enough to extract the relevant statistical information from the analysis. This is probably a consequence of the limited size (in number of parameters) of the NNs. Second, as the sampling period $\tau$ increases, two effects offset each other: on the one hand the effective training database $\mathcal{D}^{\mathsf{a}}\left(\tau, N_{\mathsf{t}}\right)$ is a better estimate of the exact database $\mathcal{D}^{\mathsf{t}}\left(\tau, N_{\mathsf{t}}\right)$ as it is less sensitive to the quality of the analysis, but on the other hand the problem of correcting a longer forecast is harder. 

%--------------------------------------------------
\subsection{Comparison between the neural networks}
\label{ssec:sec5-comparison-ml-models}
%--------------------------------------------------

Using the ensemble of NNs defined in Section~\ref{ssec:sec5-changing-database-hyperparams}, it is possible to make a comparison between the ML models. We draw the following conclusions.

The minimisation strategy presented in Section~\ref{ssec:sec5-ml-template}, and in particular the use of a large validation database, is sufficient to detect overfitting. There are some overfitting occurrences for large, linear CD models. This probably comes from the fact that CNN layers are significantly harder to train than DNN layers. In these experiments, overfitting has no impact on the results, because the training stops when the overfitting starts. However, for more complex learning problems, regularisation could be needed to counteract overfitting. 

As expected, the NNs with more nodes/filters per layer yield better forecasts in general. This is not a surprise because these models are bigger (in number of parameters). However, there is a limit to the model size which can be trained using such small databases. Indeed, we observe that the difference between \numlist{4;8} nodes/filters per layer is much larger than the difference between \numlist{8;16}. By contrast, we notice that the NNs with more layers (\textit{i.e.} the deeper models) do not yield better forecasts. This contradicts the general belief in ML that deeper networks tend to perform better. However, we think that this aspect is specific to our regression problem, and in particular to its weak nonlinearity, and that it may not generalise to other problems.

\begin{figure}[tb]
    \centering
    \includegraphics[scale=\scale]{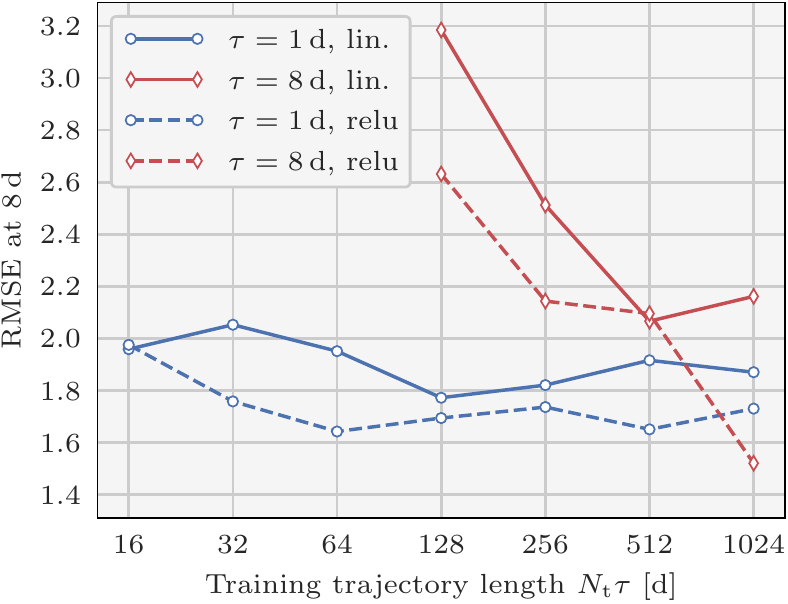}
    \caption{Evolution of the FS at $\SI{8}{\day}$ as a function of the total length of the training trajectory $N_{\mathsf{t}}\tau$ in number of days for several values of the sampling period $\tau$. For each experiment, the FS is computed using an ensemble of $N_{\mathrm{e}}=\num{16}$ initial conditions, and we select the NN which yields the lowest FS at $\SI{8}{\day}$ over all $\num{12}$ NNs with linear activation (continuous lines) or all $\num{12}$ NNs with relu activation (dashed lines).}
    \label{fig:sec5-activation-summary-8}
\end{figure}

Finally, two further points can be highlighted. First, the performance of the CD models is overall equivalent to that of the D models. This is very important because for realistic systems, in which the state space is high-dimensional, D models would be probably unaffordable. Second, the models with relu activation noticeably outperform the models with linear activation. This is illustrated in Figure~\ref{fig:sec5-activation-summary-8}, and it suggests that the nonlinearity of the regression problem, even if weak, plays a visible role.

%--------------------------------------------------
\subsection{Training with dense, noiseless observations}
%--------------------------------------------------

\begin{figure}[tb]
    \centering
    \includegraphics[scale=\scale]{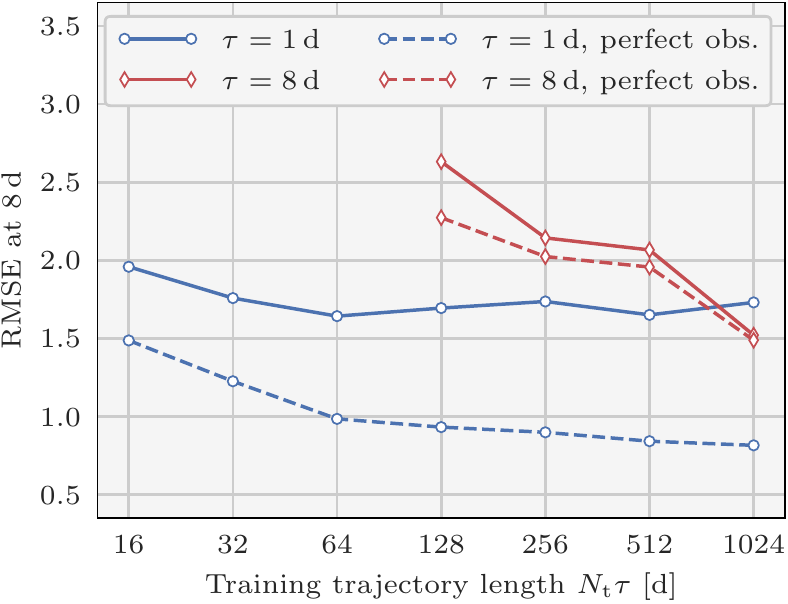}
    \caption{Evolution of the FS at $\SI{8}{\day}$ as a function of the total length of the training trajectory $N_{\mathsf{t}}\tau$ in number of days for several values of the sampling period $\tau$. For each experiment, the FS is computed using an ensemble of $N_{\mathrm{e}}=\num{16}$ initial conditions, and we select the NN which yields the lowest FS at $\SI{8}{\day}$ over all $\num{24}$ NNs trained with $\mathcal{D}^{\mathsf{a}}_{n}\left(N_{\mathsf{t}}\right)$ (continuous lines) or $\mathcal{D}^{\mathsf{t}}_{n}\left(N_{\mathsf{t}}\right)$ (dashed lines).}
    \label{fig:sec5-ideal-summary-8}
\end{figure}

To close this series of ML experiments, all $\num{24}$ NNs are trained again but using $\mathcal{D}^{\mathsf{t}}\left(\tau, N_{\mathsf{t}}\right)$ instead of $\mathcal{D}^{\mathsf{a}}\left(\tau, N_{\mathsf{t}}\right)$, in other words with dense, noiseless observations. Once again, for each training configuration, we select the NN which yields the best results over all $\num{24}$ NNs. Figure~\ref{fig:sec5-ideal-summary-8} shows the evolution of the FS at $\SI{8}{\day}$ as a function of the total length of the training trajectory $N_{\mathsf{t}}\tau$. We observe that (i) at fixed sampling period $\tau$, the FS improves as the training trajectory gets longer and (ii) globally the FS improves as the sampling period $\tau$ is increased. As mentioned in Section~\ref{ssec:sec5-changing-database-hyperparams}, this is what is typically expected for such an inference problem. Furthermore, taken into account that the NNs are small and that the training databases are short, the performance is remarkable. Finally, the most important result of this experiment is that the difference between the NNs trained with $\mathcal{D}^{\mathsf{t}}\left(\tau, N_{\mathsf{t}}\right)$ and those trained with $\mathcal{D}^{\mathsf{a}}\left(\tau, N_{\mathsf{t}}\right)$ gets smaller as the sampling period $\tau$ increases. This confirms the arguments given in Section~\ref{ssec:sec5-changing-database-hyperparams} regarding the quality of the effective training database $\mathcal{D}^{\mathsf{a}}\left(\tau, N_{\mathsf{t}}\right)$.

Furthermore, this set of idealised ML experiment has also been performed without using the original model $\mathcal{M}^{\mathsf{o}}$. In this case, the ML model $\mathcal{M}^{\mathsf{ml}}$ approximates the full model dynamics. The results (not shown here) corroborate the conjecture by \citet{jia-2019, watson-2019} mentioned in the introduction: the surrogate model is harder to train, requiring a larger training database, and eventually less accurate (as measured by the forecast skill) when learning the full model dynamics than when correcting the original model $\mathcal{M}^{\mathsf{o}}$.

%--------------------------------------------------
\section{The corrected data assimilation step}
\label{sec:sec6-corrected-da-step}
%--------------------------------------------------

As mentioned in Section~\ref{ssec:sec2-optimisation-strategy}, with a realistic model it may be computationally unaffordable to perform more than one DA--ML step, which is why we stop the algorithm here. Nevertheless, we want to evaluate the potential improvements from the correction in a subsequent DA experiment. In this section, we present the implementation and the results for this corrected DA step.

%--------------------------------------------------
\subsection{Data assimilation setup}
%--------------------------------------------------

For this series of experiments, we use the same observation database as for the original DA step (presented in Section~\ref{ssec:sec4-observation-database}). Once again, the observations are assimilated using the (strong-constraint) 4D-Var algorithm. The cost function to minimise is the same, Equation~\eqref{eq:sec4-def-cost-4dvar}, but using the resolvent of the hybrid surrogate model $\mathcal{M}^{\mathsf{o}}+\mathcal{M}^{\mathsf{ml}}$ in place of the resolvent of the original model $\mathcal{M}^{\mathsf{o}}$.

From a technical point of view, we perform the corrected DA step using the forcing formulation implementation of the weak-constraint 4D-Var algorithm \citep{laloyaux-2020}. In this formulation, the resolvent of the model is $\mathcal{M}^{\mathsf{o}}+\eta$, where $\eta$ is the forcing term, corresponding to a single integration step $\delta t$ and being by assumption constant over the DAW. In practice, for each DAW, $\eta$ is computed as
\begin{equation}
    \eta = \frac{\delta t}{\tau} \mathcal{M}^{\mathsf{ml}}\left(\mathbf{x}^{\mathsf{b}}\right),
\end{equation}
where $\mathbf{x}^{\mathsf{b}}$ is the background state of the DAW. Furthermore, the $\eta$ update in the weak-constraint 4D-Var algorithm is disabled. The major advantage with this method is that it is non-intrusive and can be implemented right away with existing tools. However, one must keep in mind that we implicitly assumed that \emph{model error grows linearly in time}, which explains the scaling factor $\delta t/\tau$. 

%--------------------------------------------------
\subsection{Data assimilation results}
\label{ssec:sec6-da-results}
%--------------------------------------------------

For simplicity, the experiments are performed with only one of the $\num{24}$ NNs. We choose to use the D model made of $\num{1}$ DNN layer with $\num{8}$ nodes and linear activation functions. This NN has obtained overall good results (compared to the other NNs) in all configurations tested in Section~\ref{ssec:sec5-changing-database-hyperparams}. Preliminary experiments have shown that the other NNs yield very similar results.

The method is applied to the $\num{16}$ test trajectories. A total of $\num{136}$ consecutive DA cycles are performed with success. The analysis RMSE stabilises after about $\num{5}$ cycles. Hence the first $\num{8}$ cycles are dropped as spinup cycles. The middle panel of Figure~\ref{fig:sec6-summary} shows the evolution of the time-averaged analysis RMSE, averaged over the $\num{16}$ test trajectories, for several values of $\tau$ and $N_{\mathsf{t}}$. 

Overall, the analysis RMSE is lower with the hybrid surrogate model than with the original model. In the best cases, the analysis RMSE reduction is about $\SI{25}{\percent}$. As for the experiments shown in Section~\ref{ssec:sec5-changing-database-hyperparams}, the size of the database $N_{\mathsf{t}}$ has little influence on the results. This confirms that only small databases are necessary to extract the relevant statistical information from the analysis. By contrast in this case, we observe that the sampling period $\tau$ has a significant influence. Even though from a theoretical perspective, the duration of the short-term linear evolution of the model error is proportional to the doubling time of errors \citet{nicolis-2003}, which is in our case (around $\SI{250}{\hour}$) much larger than the tested sampling periods, we have found that the main source of error comes from the assumption of a linear growth of model error in time. For illustration, Figure~\ref{fig:sec3-snap} shows that the model error for a $\SI{1}{\hour}$-integration is visually very different from the model error for a $\SI{1}{\day}$- or $\SI{2}{\day}$-integration. Therefore, we expect the analysis to get worse as the sampling period $\tau$ increases (\textit{i.e.} as the model error growth in time becomes significantly nonlinear), which is confirmed by Figure~\ref{fig:sec6-summary}. 

In order to better understand this phenomenon, we have performed additional experiments in idealised conditions, \textit{i.e.}, when the NN exactly predicts the model errors. In this case, it is possible to use a sampling period $\tau$ shorter than $\Delta T=\SI{1}{\day}$, the length of the DAW. In these experiments, the analysis RMSE stops improving when the sampling period $\tau$ is lower than $\SI{3}{\hour}$. We conclude that the model error growth is linear in time only during the first few hours. Moreover, for very short values of the sampling period $\tau$ (typically $\tau=\SI{1}{\hour}$), it is possible to further improve the analysis RMSE by taking into account the time evolution of the model error \emph{inside} the DAW.

Finally for comparison, using the forcing formulation of the weak-constraint 4D-Var (online model error estimation) in the exact same DA problem yields an analysis RMSE about $\num{0.07}$. This shows that there is room for improvement for our offline model error estimation method. 

%--------------------------------------------------
\subsection{Sensitivity to the observation density}
%--------------------------------------------------

\begin{figure}[tb]
    \centering
    \includegraphics[scale=\scale]{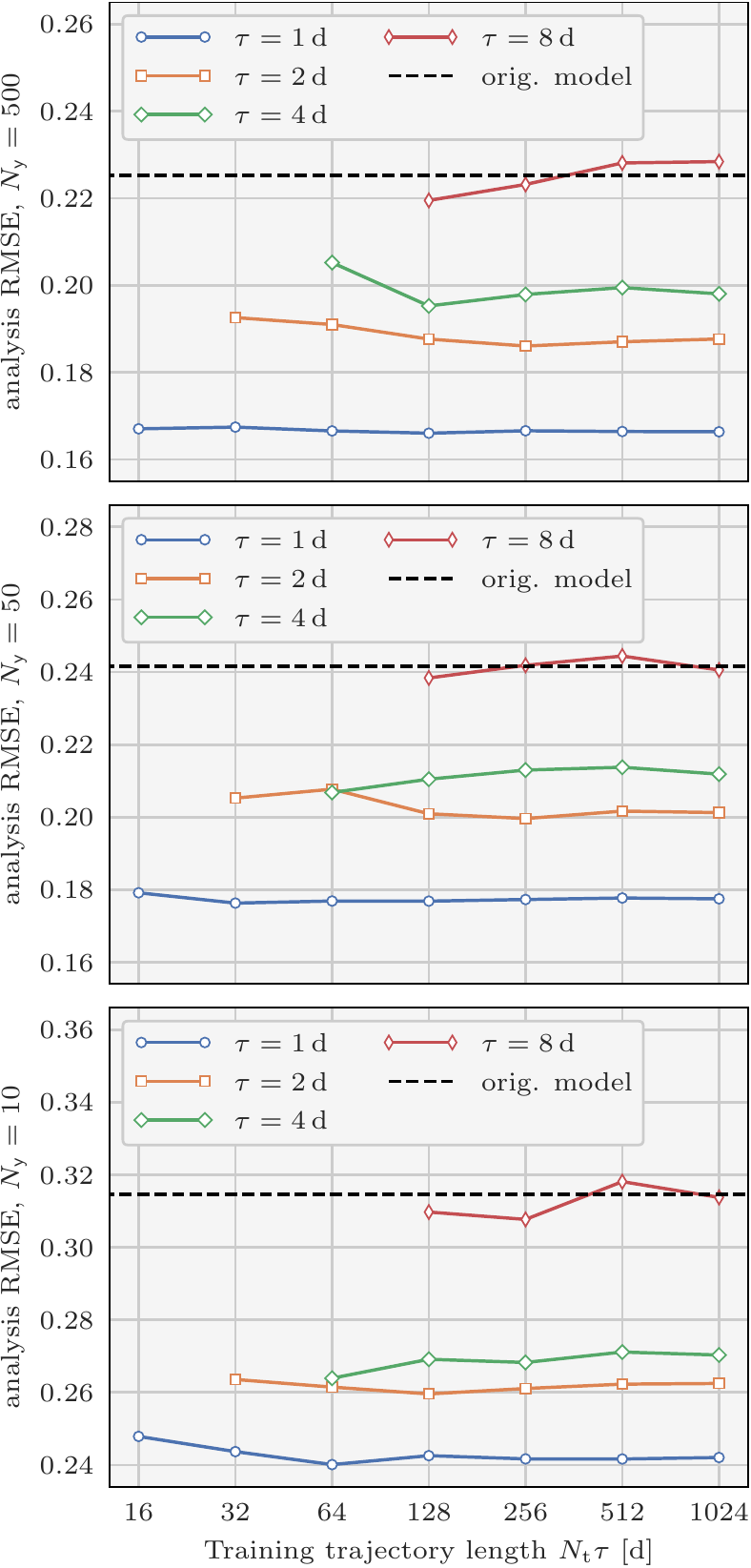}
    \caption{Evolution of the time-averaged analysis RMSE as a function of the total length of the training trajectory $N_{\mathsf{t}}\tau$ in number of days for several values of the sampling period $\tau$. The observation density $N_{\mathsf{y}}$ is set to $\num{500}$ (top panel), $\num{50}$ (middle panel), or $\num{10}$ (bottom panel) observations every $\Delta t=\SI{2}{\hour}$. For comparison, the time-averaged analysis RMSE obtained with the original model only is shown with an horizontal, dashed black line.}
    \label{fig:sec6-summary}
\end{figure}

In order to illustrate the sensitivity to the observation density, the whole set of experiments (original DA step, ML step, corrected DA step) is performed again with an observation density $N_{\mathsf{y}}$ of \numlist{10;500} observations (instead of $\num{50}$) every $\Delta t=\SI{2}{\hour}$. The implementation of the DA--ML method is exactly the same as before, except for the standard deviation $b$ of the background error covariance matrix $\mathbf{B}$, which is set to $\num{0.16}$ when $N_{\mathsf{y}}=\num{10}$ and to $\num{0.022}$ when $N_{\mathsf{y}}=\num{500}$. Figure~\ref{fig:sec6-summary} shows the results for the corrected DA step. Overall, the results are qualitatively similar to the standard case $N_{\mathsf{y}}=\num{50}$, with an analysis RMSE reduction of the same order in all three cases. It is remarkable to see that the method works even with as few observations as $N_{\mathsf{y}}=\num{10}$ every $\Delta t=\SI{2}{\hour}$, thanks to the robustness of the 4D-Var algorithm.

%--------------------------------------------------
\subsection{Further improving the analysis}cd sub
%--------------------------------------------------

To close this series of corrected DA experiments, we mention several options to further improve the analysis. As mentioned in Section~\ref{ssec:sec6-da-results}, the error growth in time has a significant nonlinear contribution when the lead time is higher than a few hours. Therefore, an efficient way to decrease the analysis RMSE would be to build a NN able to predict the error for a lead time shorter than $\Delta T=\SI{1}{\day}$, the length of the DAW. For example, using a lead time of $\SI{6}{\hour}$ should already bring a significant improvement.

The most natural and flexible way to do this is to follow the idea described in Section~\ref{sssec:sec2-correcting-flow}, that is to correct the tendencies instead of the resolvent. However, as explained in Section~\ref{sssec:sec2-correcting-flow}, this method requires substantial modifications to the existing numerical methods and it cannot be implemented without a further effort.

The other possibility is to decrease the sampling period $\tau$ to $\SI{6}{\hour}$. As is, the sampling period is bounded by $\Delta T$, the length of the DAW in the DA step, but at least two different ways can be considered to overcome this issue. First, we can use overlapping DAWs and second, we can perform several independent, cycled DA steps starting at different times of the day. In both cases, the number of 4D-Var problems to solve is multiplied by four\footnote{In the second case, it is less problematic since the independent DA steps can be performed in parallel.}. Furthermore in the first case, the observation standard deviation should be adjusted to take into account the number of times each observation is used. However, one should keep in mind that the RMS norm of the $\SI{6}{\hour}$ model error is about one fourth of the RMS norm of the $\SI{1}{\day}$ model error. This means that we need either a more efficient DA method (with a lower analysis RMSE) or more DA--ML cycles.

%--------------------------------------------------
\section{Conclusions}
\label{sec:sec7-conclusions}
%--------------------------------------------------

In this article, we have studied the possibility to use sparse and noisy observations to correct model error. The case of sparse and noisy cannot be rigorously treated with basic ML methods. To overcome this issue, we choose to combine DA and ML as originally proposed by \citet{brajard-2020}. Their method is iterative, and each iteration alternates a DA step to estimate the state from the observations, with a ML step, to learn the underlying dynamics of the DA analysis. The key output of the method is a surrogate model meant to emulate the dynamical system. Instead of constructing the surrogate model from scratch, we have proposed to build a hybrid model using an already existent (original) model. In this case, the trainable part of the hybrid model would be correcting the error of the original model, which is the primary goal in this paper. There are mainly two advantages to do so: first, we avoid the issue of a cold start (which could yield a quick divergence) in the method of \citet{brajard-2020} and second, correcting the original model is likely to be an easier learning problem than constructing the full model dynamics \citep{jia-2019, watson-2019}. In practice, two formulations can emerge from this idea, depending on whether the correction is applied to the resolvent or directly to the tendencies. Both methods have advantages and drawbacks, but in this article we chose to focus on the first one because it can be implemented without modifying the existent numerical methods.

The method has been evaluated using a two-layer, two-dimensional QG model. Model error is introduced by the means of perturbed parameters (layer depths, orography term, integration step). We have focused on the first DA--ML cycle of the model error statistical learning method. The original DA step consists in a standard DA step, in which we use the original model. It has been performed without difficulty with the strong-constraint 4D-Var algorithm. Then, in the ML step, we have used the analysis increments to train the ML models. We have first shown that a small NN is enough to predict a significant part of the model error variance, and that the resulting hybrid surrogate model has a better forecast skill than the original model up to a $\SI{16}{\day}$-forecast. In a more systematic way, we have tested different NNs as well as different training configuration (sampling period and database length). In general, bigger NNs (in number of parameters) perform better. We have also found that nonlinear NNs significantly outperform linear NNs and that, to some extent, CNN layers can be used in place of DNN layers without altering the performances. The former emphasises the nonlinearity of the learning problem, and the latter is very important because for realistic systems, ML models based on DNN layers only would be probably computationally unaffordable. Furthermore, we have concluded that short trajectories are enough to extract the relevant statistical information from the analysis during the ML step. Finally, we have illustrated how the hybrid surrogate can be used in a corrected DA step, by assuming that the error grows linearly in time. Overall, we have obtained a substantially better analysis with the hybrid surrogate model than with the original model. Furthermore, the improvement is more important when the sampling period (during the ML step) is lower, which we have explained by nonlinearities in the error growth, and we have proposed several ideas to further improve the analysis.

At this point, several topics are possible for further studies. First, we did not choose to use ML to correct the tendencies because this would have required substantial modifications to both the DA and the ML methods. However, we have explained that doing so could yield a better analysis in the corrected DA step. Implementing such a correction would confirm or infirm the idea. Second, the model error correction method presented here is by construction offline, and it would be interesting to make it online. Indeed, we have shown that the forcing formulation of the weak-constraint 4D-Var \citep{laloyaux-2020}, an online model error correction method, yields a better analysis. Our experiments show that it is possible to use a very short database (only $\num{16}$ samples) without altering the results much, which tends to confirm the possibility of an online learning. However, one must keep in mind that, in our experimental setup we have $\num{600}$ observations per DAW. This is enough to learn the state at the start of the window ($\num{1600}$ variables), but it is unclear whether it will be enough to train in addition a ML model with more than $\num{E4}$ parameters.

\section*{acknowledgements}
A. Farchi has benefited from a visiting grant of the ECMWF. CEREA is a member of Institut Pierre-Simon Laplace (IPSL). Furthermore, the author would like to thank three anonymous reviewers for their insightful comments and suggestions which helped improving the article.

\bibliography{bibtex}

\end{document}